%% file: main.tex
\documentclass[10pt,twocolumn,letterpaper]{article}

\usepackage{cvpr}              

\input{preamble}

\definecolor{cvprblue}{rgb}{0.21,0.49,0.74}
\usepackage[pagebackref,breaklinks,colorlinks,citecolor=cvprblue]{hyperref}

\usepackage{soul}
\usepackage{hhline}

\usepackage[accsupp]{axessibility} 

\newcommand{\kk}[1]{\textcolor[rgb]{0.,0.,0.}{#1}}
\newcommand{\yy}[1]{\textcolor[rgb]{0., 0, 0.}{#1}}
\newcommand{\ryn}[1]{\textcolor[rgb]{0,0,0}{#1}}
\newcommand{\rynq}[1]{\textcolor[rgb]{0, 0, 0}{#1}}

\newcommand{\yycr}[1]{\textcolor[rgb]{0, 0, 0}{#1}}
\newcommand{\email}[1]{\hypersetup{urlcolor=black}%
    \href{mailto:#1}{#1}\hypersetup{urlcolor=blue}}
\setstcolor{blue}

\def\paperID{156} 
\def\confName{CVPR}
\def\confYear{2024}

\title{Color Shift Estimation-and-Correction for Image Enhancement}

\author{Yiyu Li\qquad Ke Xu\qquad Gerhard Petrus Hancke\qquad Rynson W.H. Lau \\
City University of Hong Kong\\
{\tt\small \email{yiyuli.cs@my.cityu.edu.hk}, \email{kkangwing@gmail.com}, \{gp.hancke,  rynson.lau\}@cityu.edu.hk}
}

\begin{document}
\maketitle
\input{sec/0_abstract}    
\input{sec/1_intro}

\input{sec/2_relatedworks}

\input{sec/3_methods}

\input{sec/4_experiments}

\input{sec/6_conclusion}

{
    \small
    \bibliographystyle{ieeenat_fullname}
    \bibliography{main}
}


\end{document}

%% file: preamble.tex
\usepackage[dvipsnames]{xcolor}


%% file: sec/0_abstract.tex
\begin{abstract}
Images captured under sub-optimal illumination conditions may contain both over- and under-exposures.
Current approaches mainly focus on adjusting image brightness, which may exacerbate color tone distortion in under-exposed areas and fail to restore accurate colors in over-exposed regions.
We observe that \ryn{over- and under-exposed} regions display opposite color tone distribution shifts, which may not be easily normalized in joint modeling as they usually do not have ``normal-exposed'' regions/pixels as reference.
%
In this paper, we propose a novel method to enhance images with both over- and under-exposures by learning to estimate and correct such color shifts.
Specifically, we first derive the color feature maps of the brightened and darkened versions of the input image via a UNet-based network, \ryn{followed by} a pseudo-normal feature generator to produce pseudo-normal color feature maps.
We then propose a novel COlor Shift Estimation (COSE) module to estimate the color shifts between the derived brightened (or darkened) color feature maps and the pseudo-normal color feature maps.
The COSE module corrects the estimated color shifts of the over- and under-exposed regions separately.
We further propose a novel COlor MOdulation (COMO) module to modulate the separately corrected colors in the over- and under-exposed regions to produce the enhanced image.
Comprehensive experiments show that our method outperforms existing approaches. \yycr{\ryn{Project webpage}: {\small \url{https://github.com/yiyulics/CSEC}}.}
\vspace{-3mm}
\end{abstract}

%% file: sec/1_intro.tex
\section{Introduction}
\label{sec:intro}

\begin{figure}[t]
\renewcommand{\tabcolsep}{0.8pt}
\begin{center}

\begin{tabular}{ccccc}

    \includegraphics[width=0.19\linewidth]{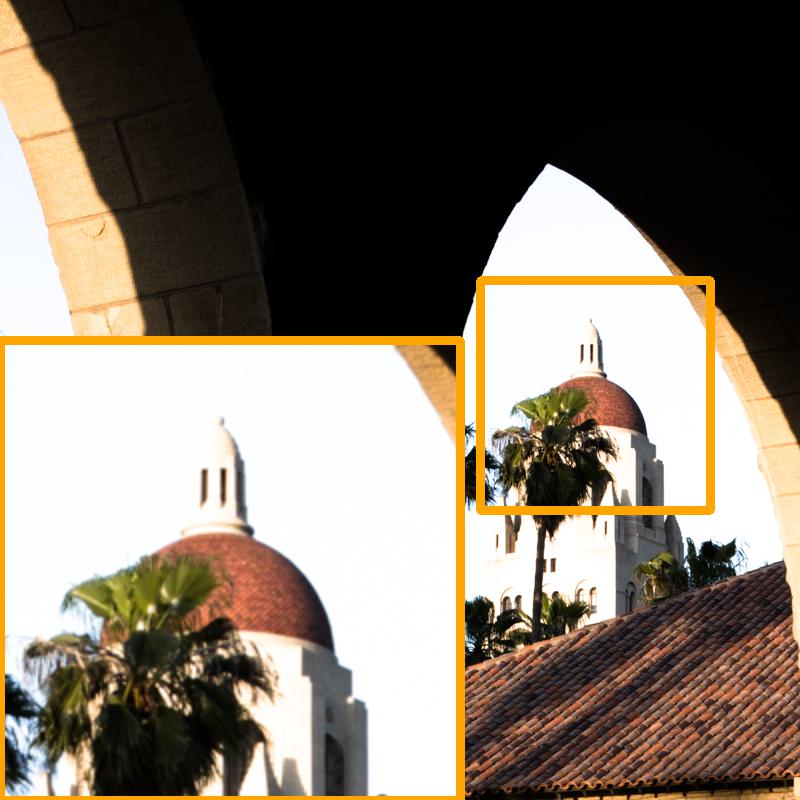} &
    \includegraphics[width=0.19\linewidth]{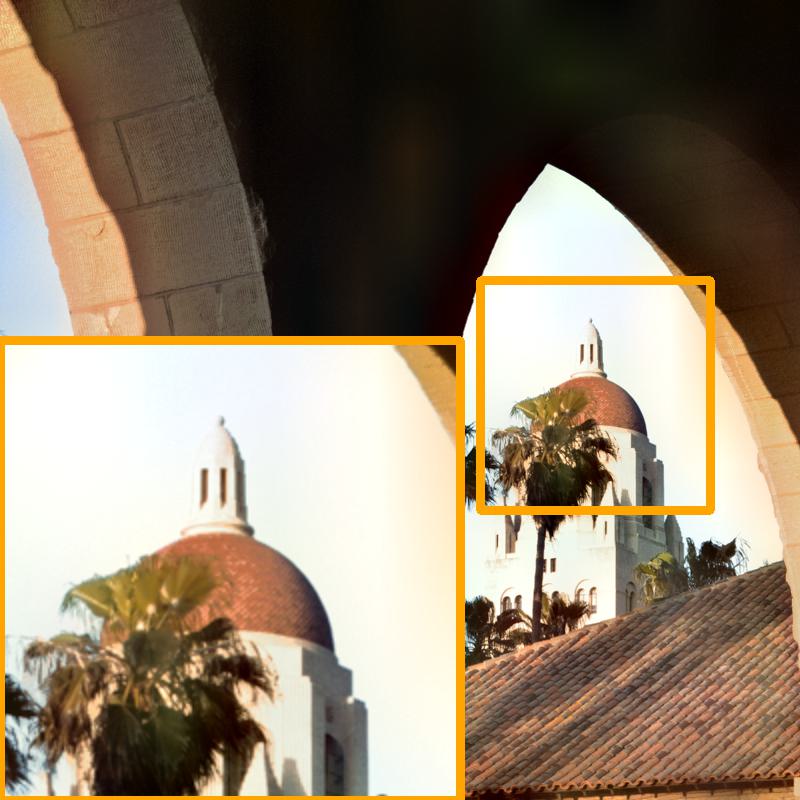} &
    \includegraphics[width=0.19\linewidth]{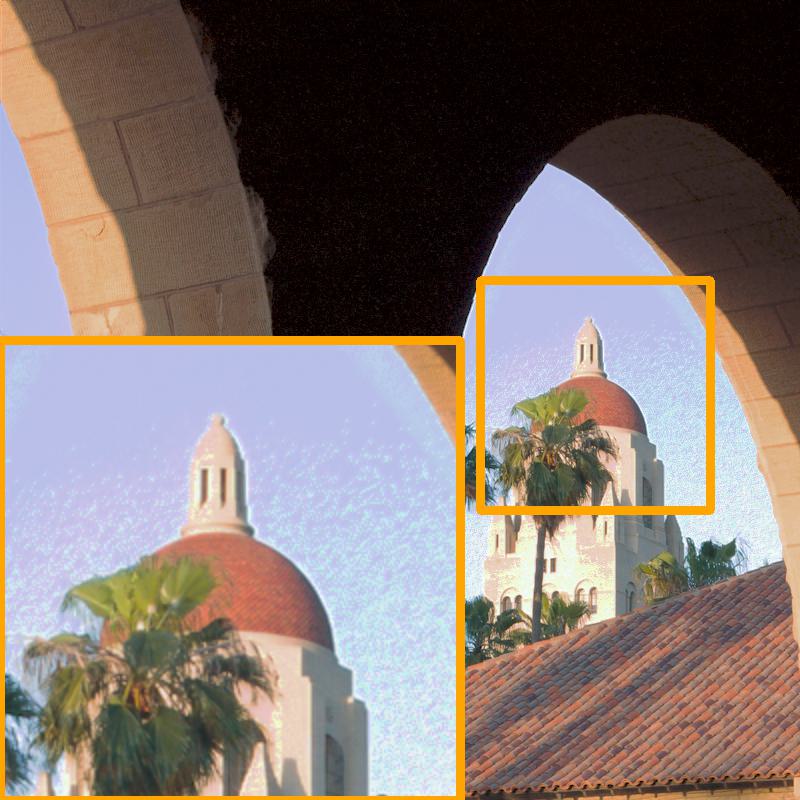} &
    \includegraphics[width=0.19\linewidth]{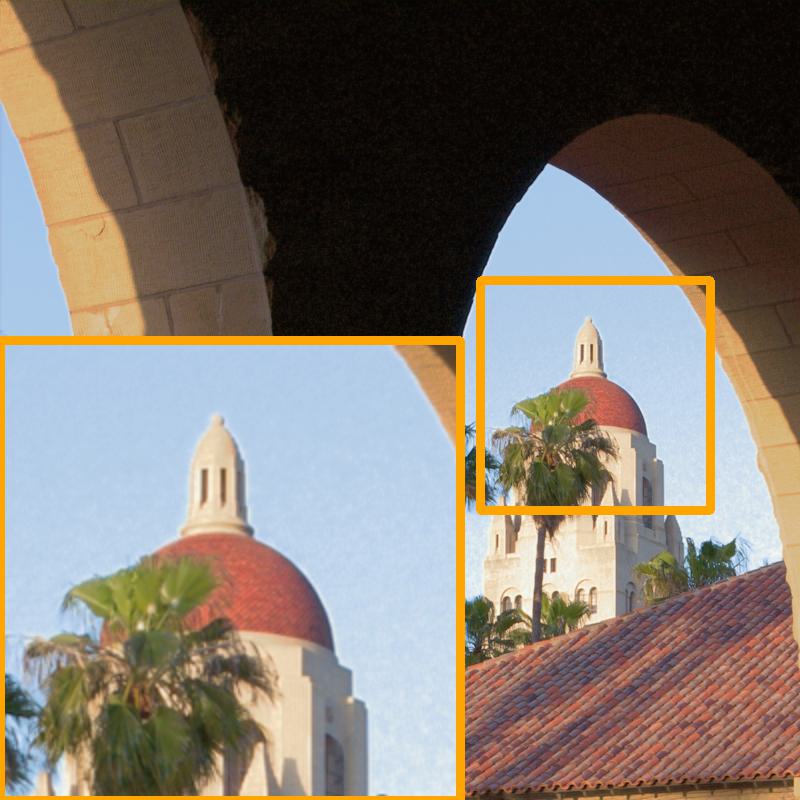} &
    \includegraphics[width=0.19\linewidth]{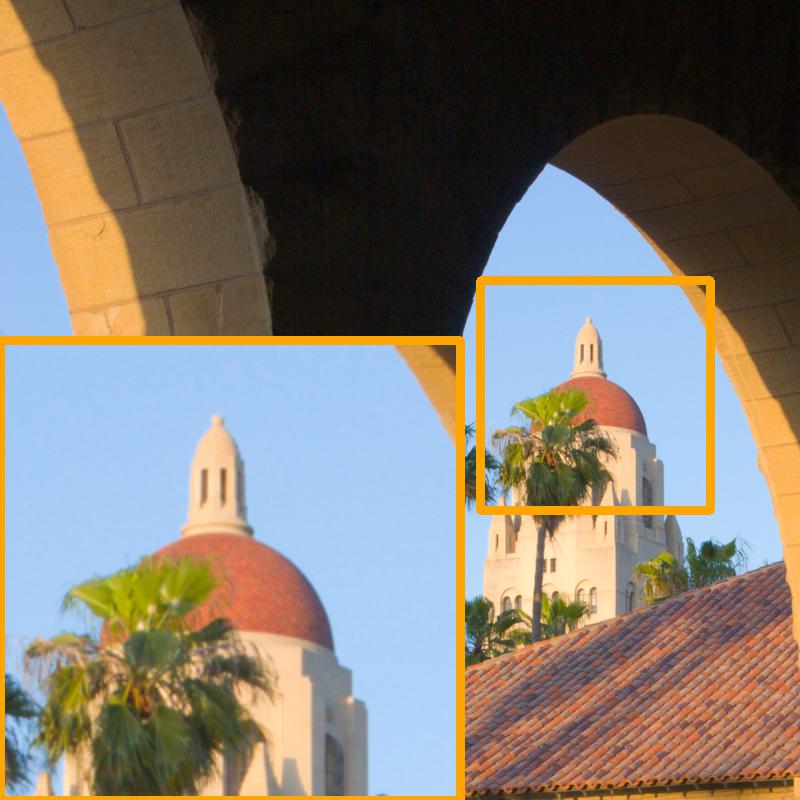}\\ 

    \includegraphics[width=0.19\linewidth]{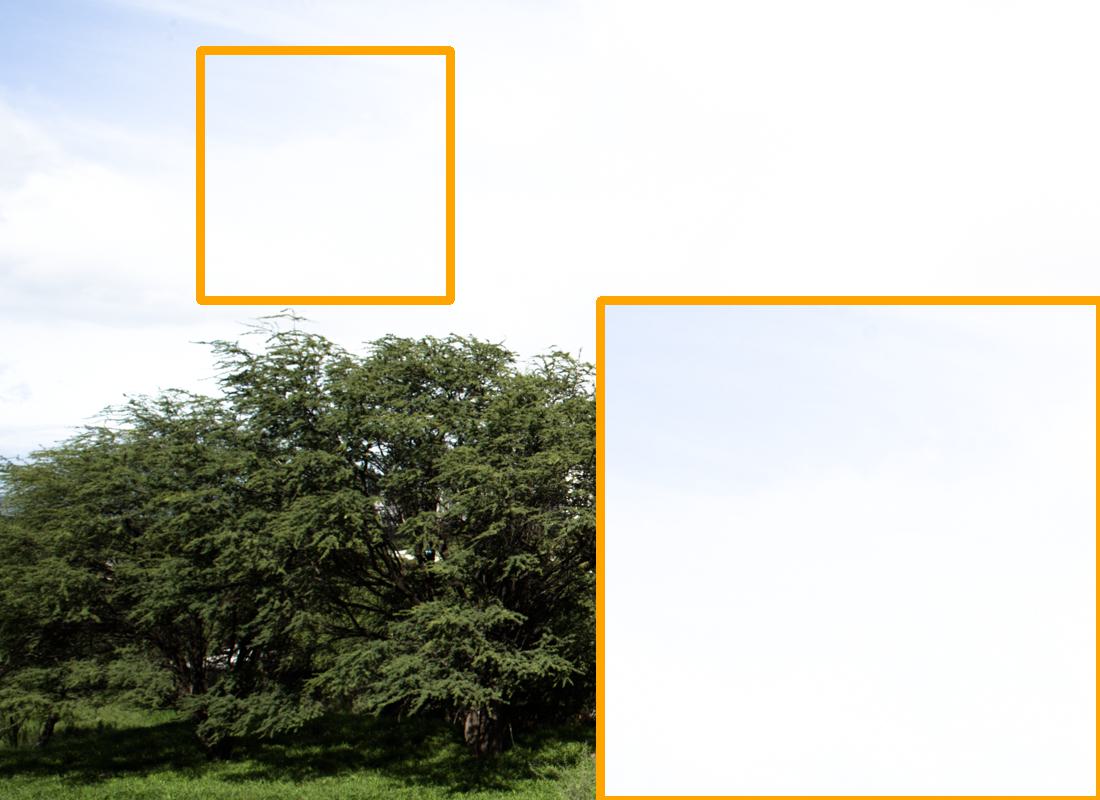} &
    \includegraphics[width=0.19\linewidth]{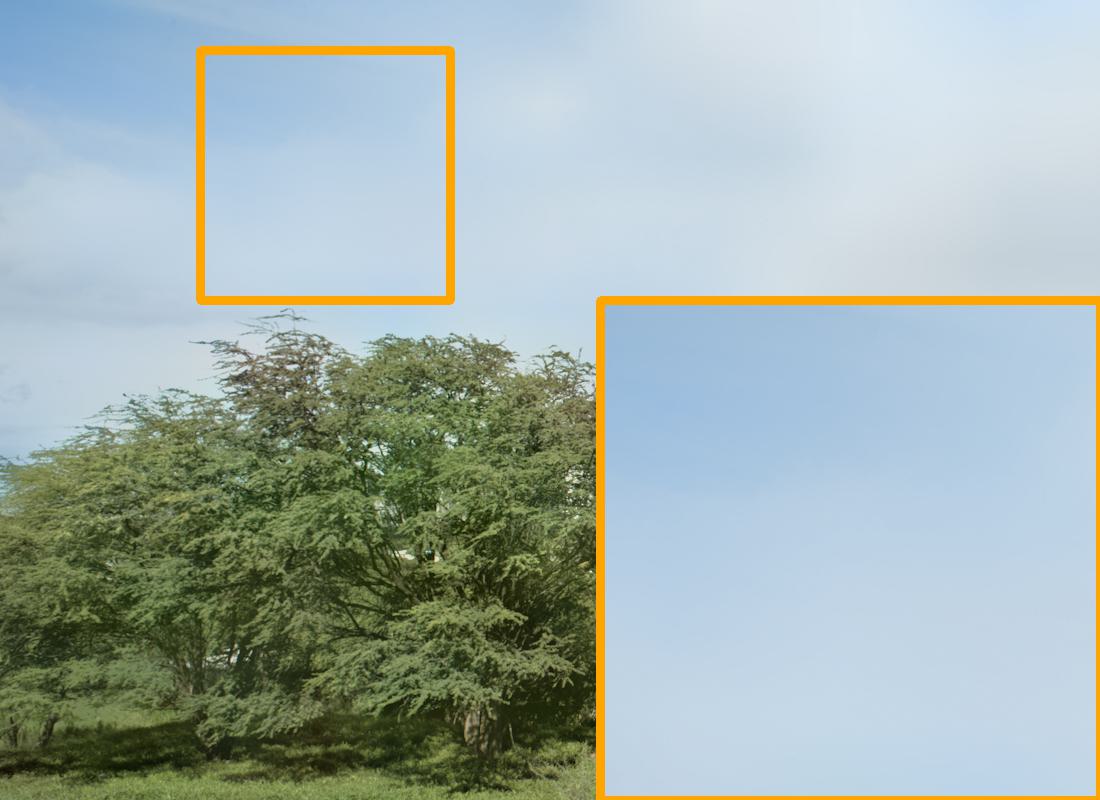} &
    \includegraphics[width=0.19\linewidth]{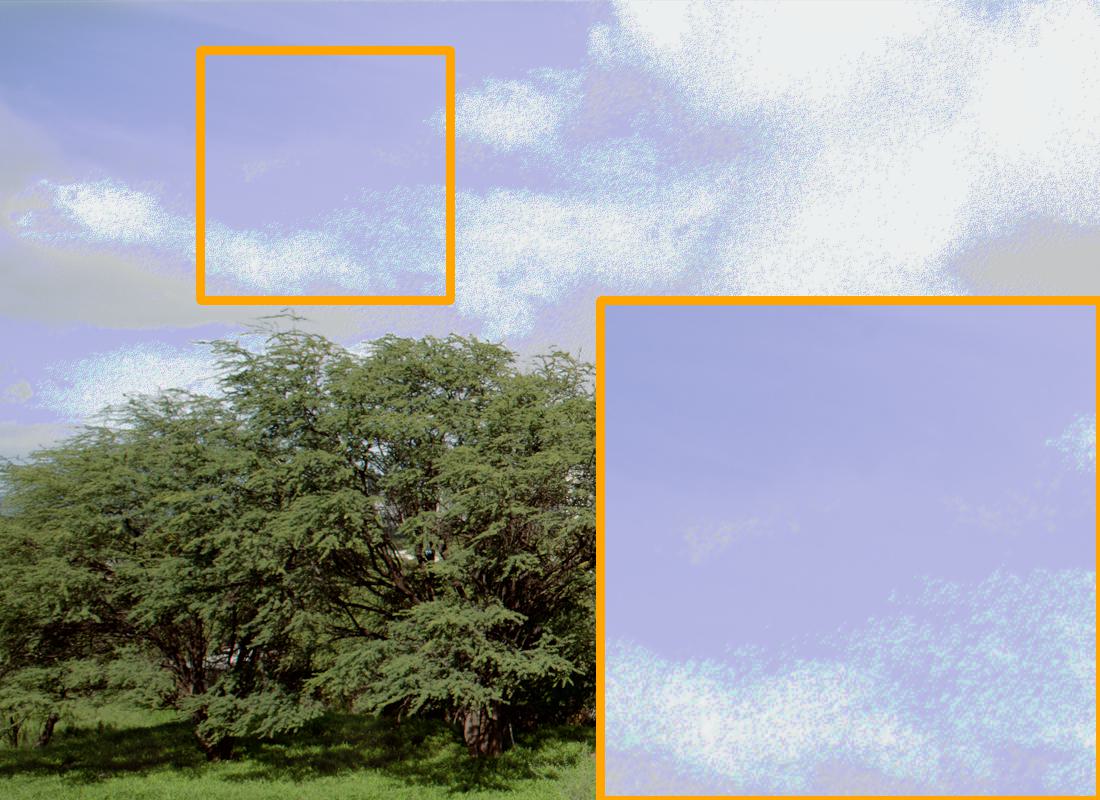} &
    \includegraphics[width=0.19\linewidth]{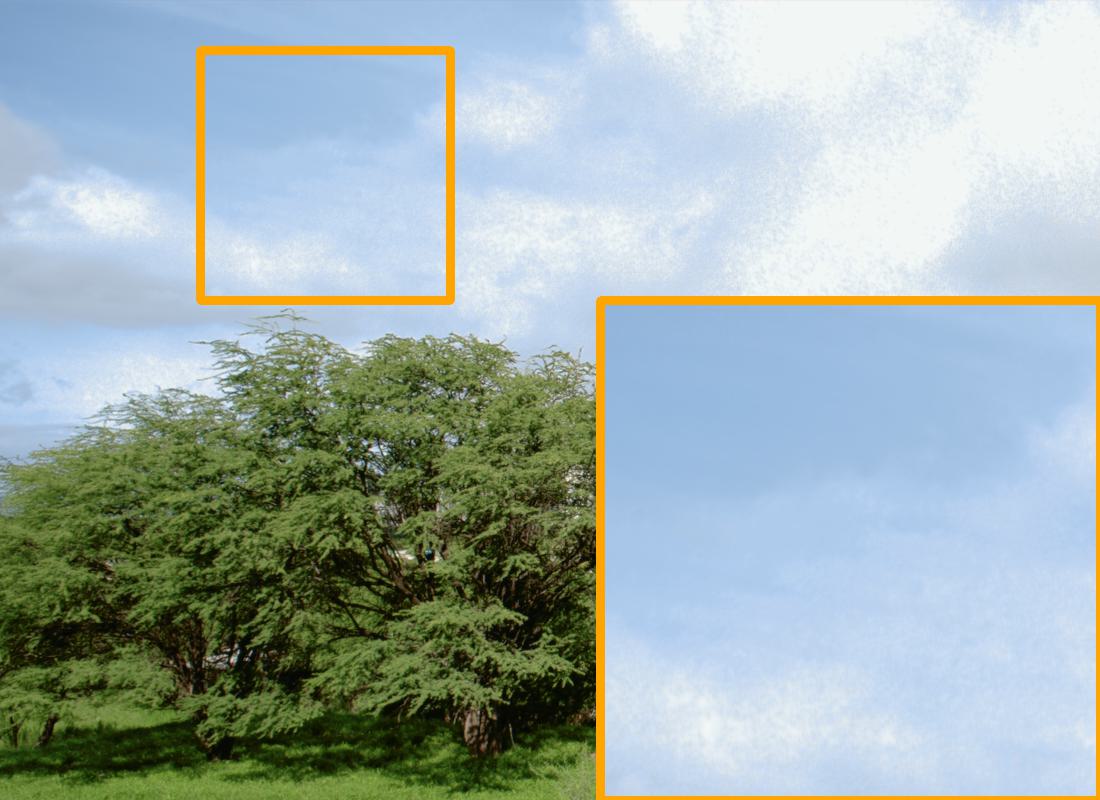} &
    \includegraphics[width=0.19\linewidth]{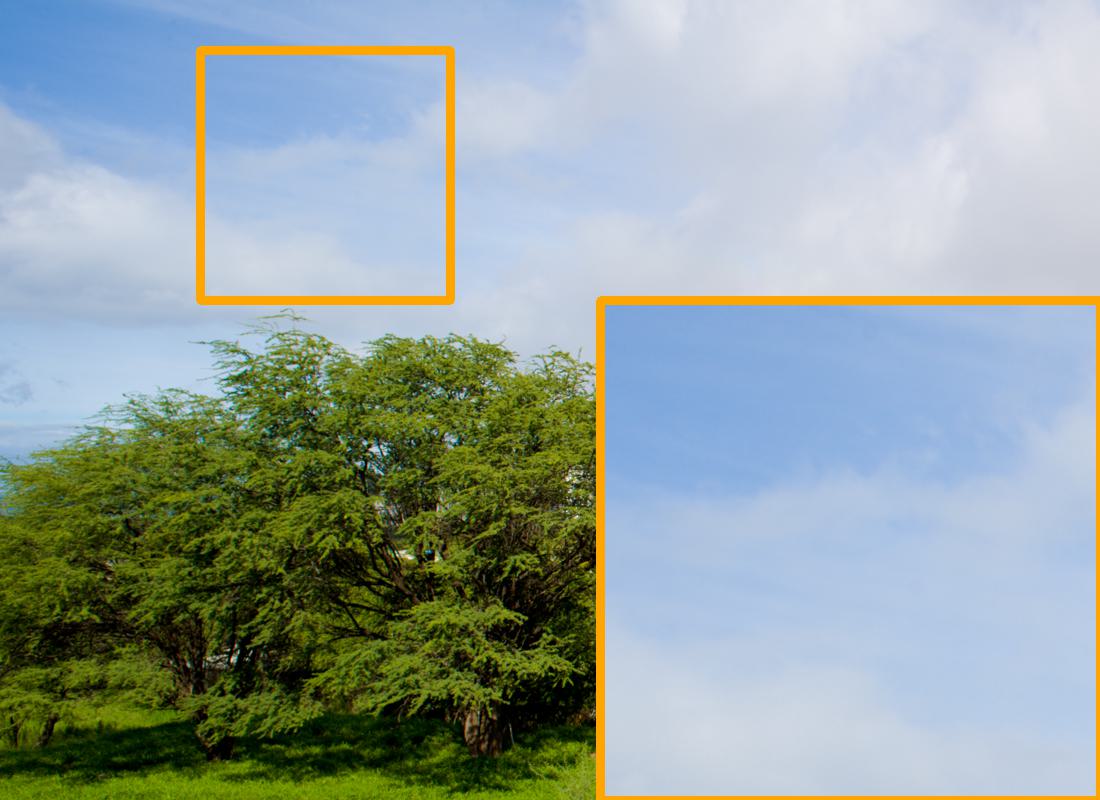}\\ 

    \includegraphics[width=0.19\linewidth]{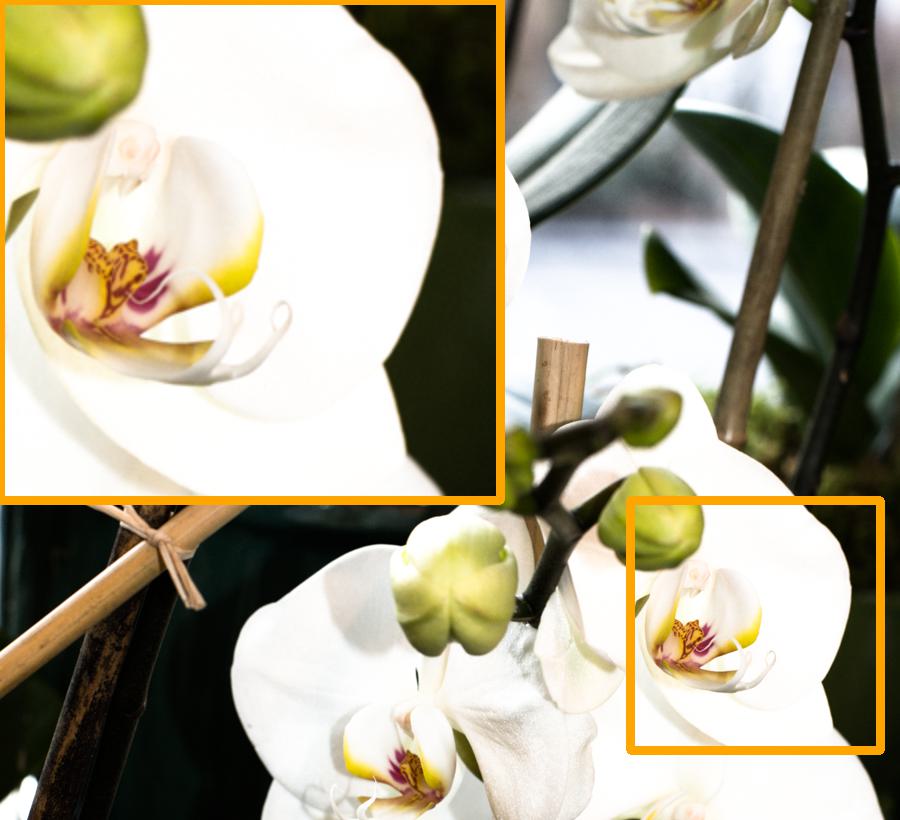} &
    \includegraphics[width=0.19\linewidth]{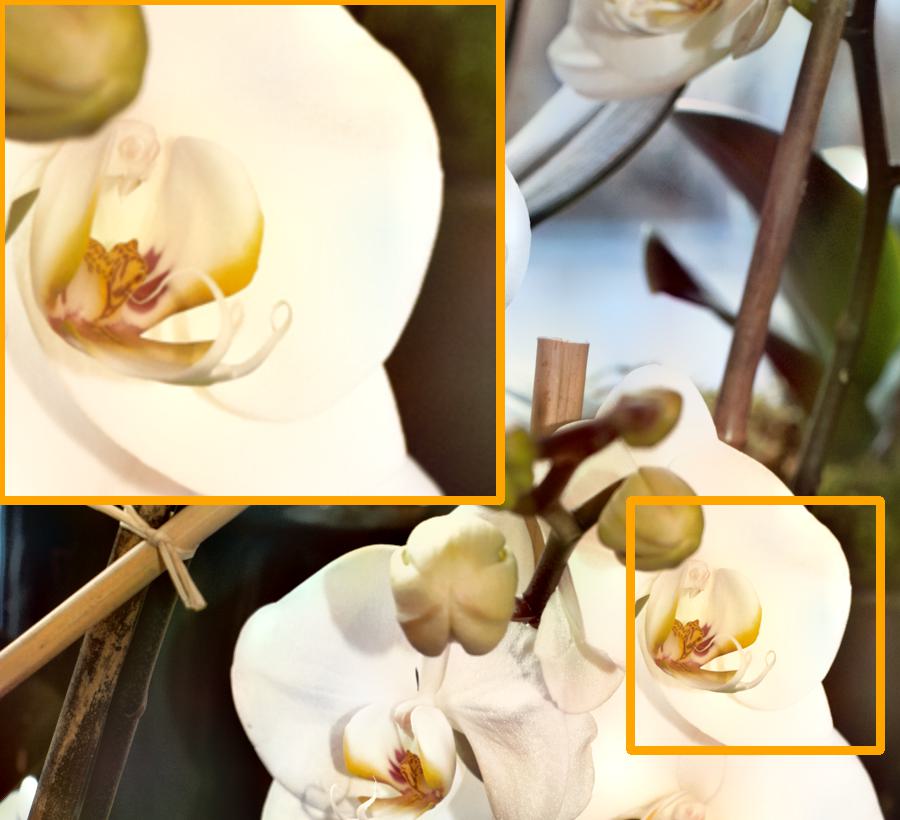} &
    \includegraphics[width=0.19\linewidth]{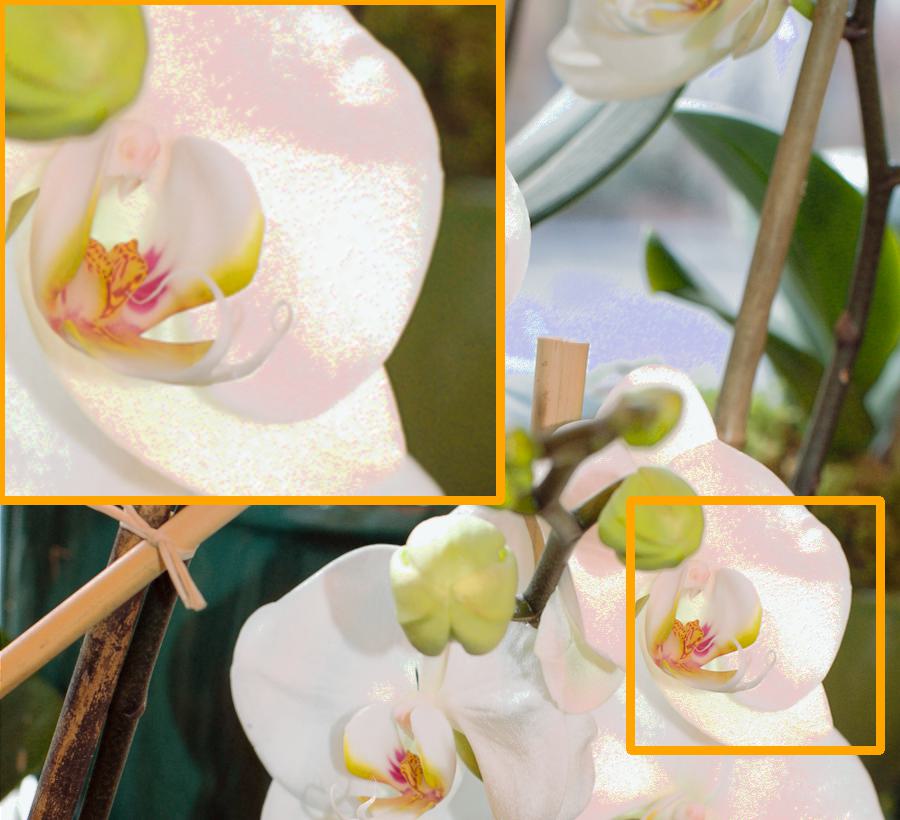} &
    \includegraphics[width=0.19\linewidth]{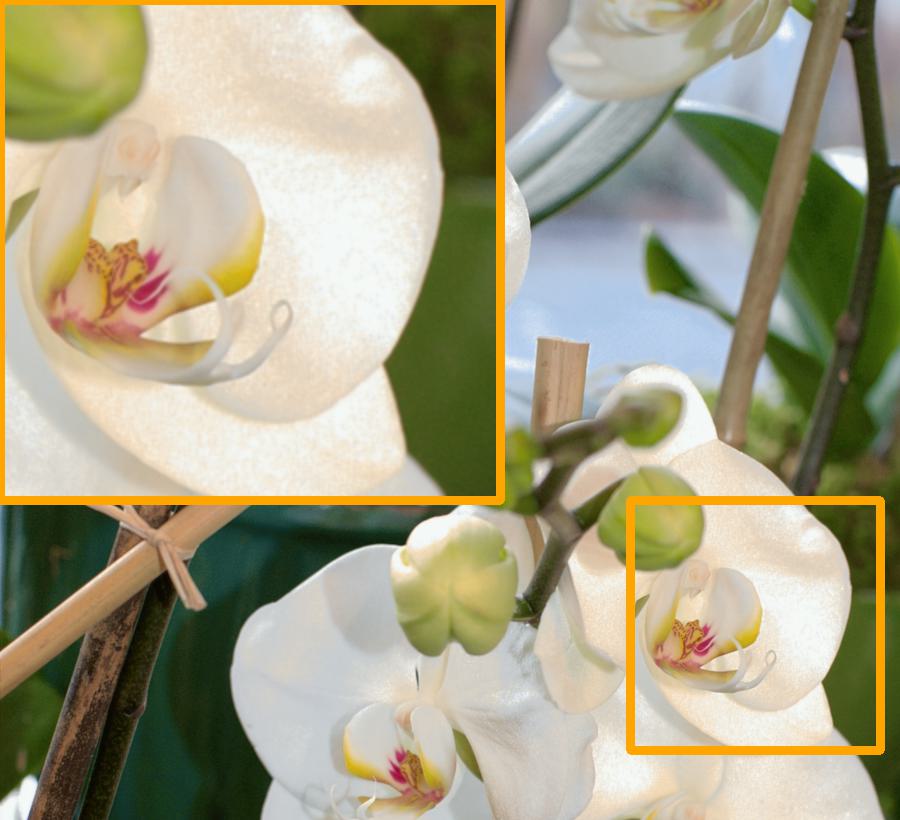} &
    \includegraphics[width=0.19\linewidth]{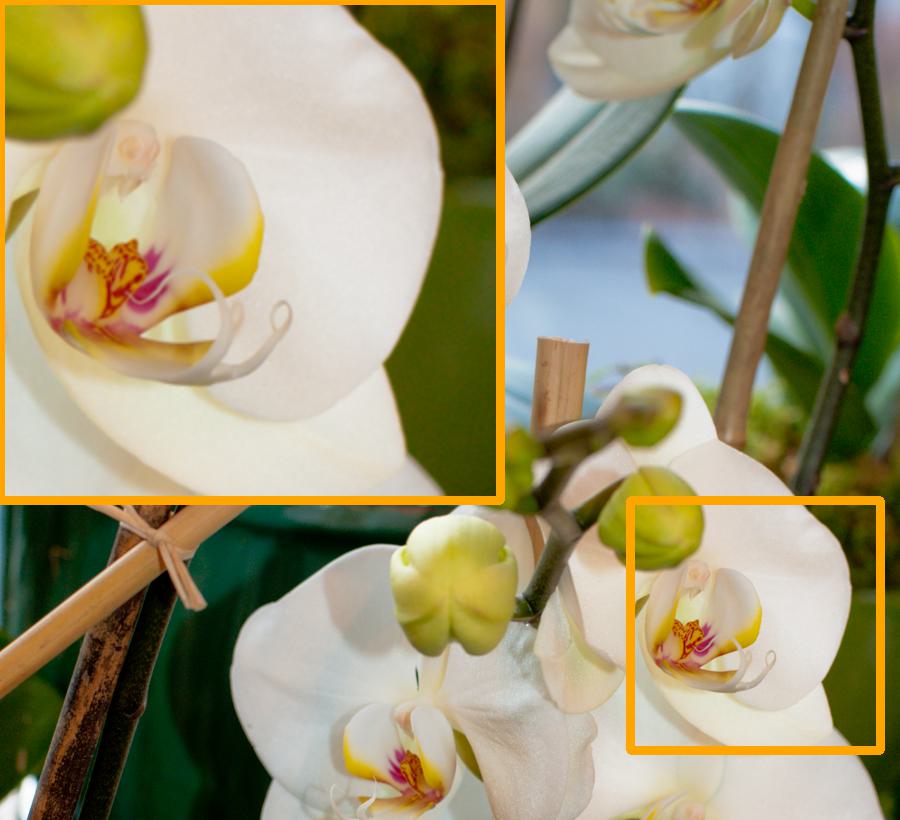}\\ 
    \scriptsize{(a) Input} & \scriptsize{(b) MSEC~\cite{Afifi_2021_CVPR_MSEC}}
    &\scriptsize{(c) LCDP~\cite{Wang_2022_ECCV_lcdp}} & \scriptsize{(d) Ours} & \scriptsize{(e) GT}\\ 


\end{tabular}

\includegraphics[width=\linewidth]{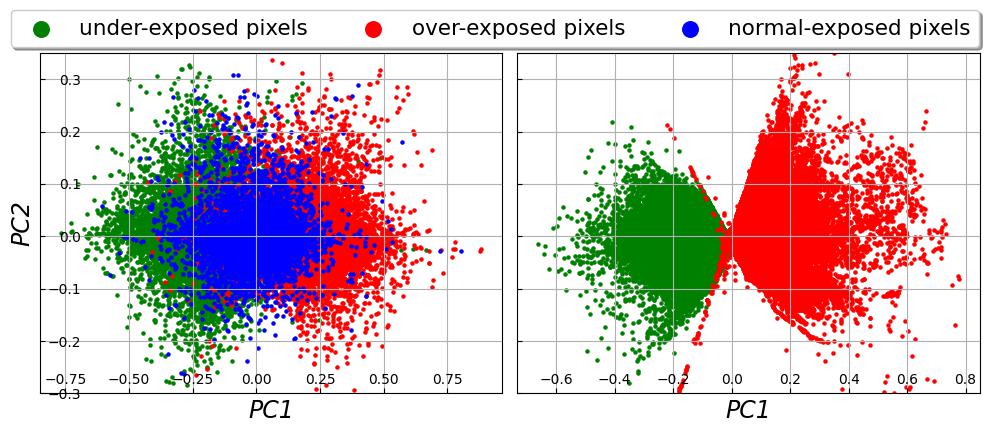}
\begin{tabular}{cc}
    \scriptsize{(f)\quad Sampled pixels in MSEC~\cite{Afifi_2021_CVPR_MSEC}\quad} & \scriptsize{\quad (g) Sampled pixels in LCDP~\cite{Wang_2022_ECCV_lcdp}\quad}\\
\end{tabular}
\end{center}
\vspace{-5mm}
\caption{Given images with both over- and under-exposures (a), \ryn{SOTA} methods~\cite{Afifi_2021_CVPR_MSEC,Wang_2022_ECCV_lcdp} may still fail to correct \ryn{color distortion} (b,c). We \ryn{show} PCA results of pixels randomly sampled from MSEC~\cite{Afifi_2021_CVPR_MSEC} (images \ryn{with} either over- or under-exposed) (f) and LCDP~\cite{Wang_2022_ECCV_lcdp} (images \ryn{with} both over- and under-exposures) (g) datasets, and make two observations.
First, we observe that under-exposed pixels (greenish dots) tend to have a reverse distribution shift compared to over-exposed pixels (reddish dots) in both datasets.
Second, unlike MSEC~\cite{Afifi_2021_CVPR_MSEC} that contains normal-exposed pixels (bluish dots) in their 0 EV input images, images with both over- and under-exposures do not have such ``reference pixels'' as guidance.
These two observations inspire us to estimate and correct such color shifts conditioned on the created pseudo-normal exposed features. Our method can properly adjust images with both over- and under-exposures (d).
}
\vspace{-5mm}
\label{fig:teaser}
\end{figure}

Real-world scenarios often involve a wide range of illumination, which poses a significant challenge for photographing.
\yy{Uneven scene illumination may easily produce both over- and under-exposures.}
Although cameras have an automatic exposure mode to determine an ``ideal'' exposure setting based on scene brightness, uniformly adjusting the exposure across the entire image may still result in excessively bright and excessively dark regions.
Such under- and over-exposed regions can exhibit obvious color tone distortions. 
The relatively high noise level in under-exposed regions alters the data distribution to cause shifts in color tone, while the saturated over-exposed regions lose the original colors.
Hence, enhancing such images typically involves brightness adjustment and color tone shift correction.

In recent years, numerous endeavors have been made to enhance images \ryn{that are improperly exposed. They} can be broadly grouped into two categories.
The first category focuses on enhancing either over-exposed or under-exposed images.
Some methods~\cite{Nsampi_2018_BMVC_CMEC, Huang_2022_CVPR_ENC} propose to learn an exposure invariant representation space, where different exposure levels can be mapped to a standardized and invariant representation.
Other methods~\cite{Huang_2022_ECCV_Fourier, Wang_2023_CVPR_Decoupling} propose to integrate frequency information with spatial information, which helps model the inherent image structural characteristics to enhance the image brightness and structure distortions.
However, as the above methods typically assume that the over- or under-exposure happens to the whole image, they do not perform well for images with both over- and under-exposures (\eg, Figure~\ref{fig:teaser}(b)).
Recently, Wang~\etal~\cite{Wang_2022_ECCV_lcdp} propose the second category of work, which aims to enhance images with both over- and under-exposures.
They utilize the local color distributions as a prior to guide the enhancement process.
However, despite their pyramidal design of local color distributions prior, their method still tends to produce results with significant color shifts, particularly in large homogeneous regions (\eg, Figure~\ref{fig:teaser}(c)).

In this paper, we aim to correct the image brightness and color distortions for images with both over- and under-exposures.
To approach this problem, we first visualize the PCA results of pixels randomly sampled from two relevant datasets, MSEC~\cite{Afifi_2021_CVPR_MSEC} (each scene has five input images of different exposure values (EV)) and LCDP~\cite{Wang_2022_ECCV_lcdp} (each scene has only one input image with both over- and under-exposures), in Figure~\ref{fig:teaser}(f) and~\ref{fig:teaser}(g), respectively.
We have two observations from this initial study. First, under-exposed pixels (green dots) tend to have an opposite distribution shift compared to over-exposed pixels (red dots) in both datasets.
Second, unlike MSEC~\cite{Afifi_2021_CVPR_MSEC} that contains 0 EV input images, which can serve as reference images for the exposure normalization process, images of LCDP~\cite{Wang_2022_ECCV_lcdp} do not contain such ``normal-exposed'' pixels.
The first observation inspires us to consider estimating and correcting \yy{such color shifts, while the second observation inspires us to create pseudo-normal exposed feature maps to function as a reference for color shift estimation and rectification.}



To this end, we propose a novel method to adjust the image brightness and correct the color tone distortion jointly.
Our method first uses a UNet-based network to derive color feature maps of the over- and under-exposed regions from the brightened and darkened versions of the input image. A pseudo-normal feature generator then creates the pseudo-normal color feature maps based on the derived color feature maps.
%
A novel COlor Shift Estimation (COSE) module is then proposed to estimate and correct the color shifts between the derived brightened (or darkened) color feature maps and the created pseudo-normal color feature maps separately. We implement the COSE module by extending the deformable convolution in the spatial domain to the color feature domain.
We further propose a novel COlor MOdulation (COMO) module to modulate the separately corrected colors of the over- and under-exposed regions to produce the enhanced image. We implement this COMO module via a tailored cross-attention mechanism performed on the input image and estimated darken/brighten color offsets.
Figure~\ref{fig:teaser}(d) shows that our method can produce visually pleasing images.

Our main contributions can be summarized as follows:
\begin{enumerate}
    \item We propose a novel neural approach to enhance images with both over- and under-exposures via modeling the color distribution shifts.
    \item We propose a novel neural network with two novel modules, a novel COlor Shift Estimation (COSE) module for the estimation and correction of colors in over- and under-exposed regions separately, and a novel COlor MOdulation (COMO) module for modulating the corrected colors to produce the enhanced images.
    \item Extensive experiments show that our network is lightweight and outperforms existing image enhancement methods on popular benchmarks.
\end{enumerate}

%% file: sec/2_relatedworks.tex
\section{Related Works}
\label{sec:relatedworks}

\noindent{\bf Exposure Correction.} When photographing scenes with complex illuminations, the \ryn{captured} images may have over- and/or under-exposure problems. Exposure correction methods~\cite{yu_nips18_deepexposure, Yang_cvpr18_drht, Zhang_2019_dual, Hu_siggraph18_white_box, Lu_eccv12_autoexposure} aim to enhance such images \ryn{to recover image details buried in over-/under-exposed} regions.
%

A group of methods focus on under-exposed image enhancement. Some of these methods are Retinex-based, which employ either handcrafted features~\cite{guo_2016_TIP_lime, fu_2016_CVPR_weighted, cai_2017_ICCV_joint} or deep features~\cite{wei_2018_BMVC_retinexnet, wang_2019_CVPR_deepupe, zhang_2019_MM_KinD, liu_2021_CVPR_ruas, zhang_2021_IJCV_KinD++}, to decompose the image into illumination and reflectance components for their enhancement.
Recently, Fu~\etal~\cite{Fu_2023_CVPR_NoPrior} propose to combine contrastive learning and self-knowledge distillation to learn the decomposition process. Fu~\etal~\cite{Fu_2023_CVPR_Instance} propose an unsupervised approach to learn the decomposition process, which uses paired images as input to guide the decomposition of each other with the consistency regularization between the two decomposed reflectance components \yy{of each image in the pair}.
Other methods directly learn the mappings from under-exposed images to normal-exposed images. These methods focus on designing various priors, \eg, frequency information~\cite{xu2020learning}, lagrange multiplier \cite{zheng_2021_cvpr_adaptive}, de-bayer filter \cite{dong_2022_CVPR_debayer}, 3D lookup table \cite{yang_2022_cvpr_adaint}, normalizing flow \cite{wang_2022_aaai_normalizing_flow}, and signal-to-noise ratio \cite{Xu_2022_CVPR_SNR}. 
Most recently, Wu~\etal~\cite{Wu_2023_CVPR_Semantic} propose to use semantic segmentation maps to help maintain color consistency. Xu~\etal~\cite{Xu_2023_CVPR_Structure} leverage the edge detection for structural modeling, which helps enhance the appearance. Wang~\etal~\cite{Wang_iccv23_llnerf} propose to reconstruct an implicit neural radiance field (NeRF) using low-light images and then enhance the NeRF to produce normal-light images.

Recently, Afifi~\etal~\cite{Afifi_2021_CVPR_MSEC} construct a large-scale dataset that contains over-, normal- and under-exposed images, and propose a Laplacian pyramid-based network for exposure correction.
Huang~\etal~\cite{Huang_2022_CVPR_ENC} propose to learn an exposure invariant space to bridge the gap between over-exposure and under-exposure features \yy{in image-level}. 
Later, Huang~\etal~\cite{Huang_2022_ECCV_Fourier} further leverage Fourier transform to combine spatial and frequency information to enhance image brightness and structures.
Most recently, Huang~\etal~\cite{Huang_2023_CVPR_Exposure} model the relationship between over-exposed and under-exposed samples in mini-batches to maintain a stable optimization of exposure correction.
Wang~\etal~\cite{Wang_2023_CVPR_Decoupling} propose to decouple the high and low frequencies of images via specially designed convolution kernels to improve the image structural modeling during exposure correction.
\yy{Yang~\etal~\cite{yang_2023_lanet} propose a unified frequency decomposition method for multiple illumination-related tasks, in which the exposure information is adjusted in their low-frequency component.}
\yy{Baek~\etal~\cite{Baek_2023_ICCV_luminance} propose a luminance-aware method for multi-exposure correction.}

\yy{The aforementioned methods focus on modeling the \ryn{exposure correction problem}
at the image-level, assuming that each image is either over- or under-exposed. \ryn{They are unable to handle}} images with both over- and under-exposures.
In contrast, Wang~\etal~\cite{Wang_2022_ECCV_lcdp} propose to leverage local color distributions to guide the network to locate and enhance  over- and under-exposed regions.
Despite the success, their
\kk{local color distributions prior often fails to correct colors, especially in large over- and under-exposed regions.} In contrast, in this work, we propose a novel approach to \ryn{address images with both over- and under-exposures through} adaptive brightness adjustment and color shift correction.

\noindent \textbf{Deformable Convolutions.}\quad Convolutional neural networks~\cite{lecun_1998_CNN} have an inherent limitation in modeling the geometric transformations due to the fixed kernel configuration. To facilitate the transformation modeling capability of CNNs, Dai~\etal~\cite{dai_2017_deformable} propose the deformable convolution operation, which adds additional spatial offsets and learns the spatial offsets from the target tasks. Later, Zhu~\etal~\cite{zhu_2019_deformablev2} propose an improved version of deformable convolution, which incorporates modulation scalar to measure the importance of different locations.
Wang~\etal~\cite{Wang_2023_deformablev3} propose to improve the deformable convolution by introducing a multi-group mechanism and sharing weights among convolutional neurons to make the model computationally efficient.
Numerous works have adopted deformable convolutions in recent years for different high-level vision tasks, \eg, image classification~\cite{Wang_2023_deformablev3}, object detection~\cite{dai_2017_deformable}, video object detection~\cite{bertasius_2018_deformable_vobject}, semantic segmentation~\cite{dai_2017_deformable}, and human pose estimation~\cite{sun_2018_deformable_humanpose}. 
Meanwhile, deformable convolution has also been adopted for low-level vision tasks, \eg, for aligning multi-frame information with a video~\cite{tian_2020_tdan, wang_2019_edvr, dudhane_2022_burst}.

While these methods typically use deformable convolutions to aggregate spatial contextual info, we propose in this work to leverage deformable convolutions to estimate color shifts of over-/under-exposed regions for enhancement.

%% file: sec/3_methods.tex
\begin{figure*}[ht!]
    \centering
    \includegraphics[width=0.85\linewidth]{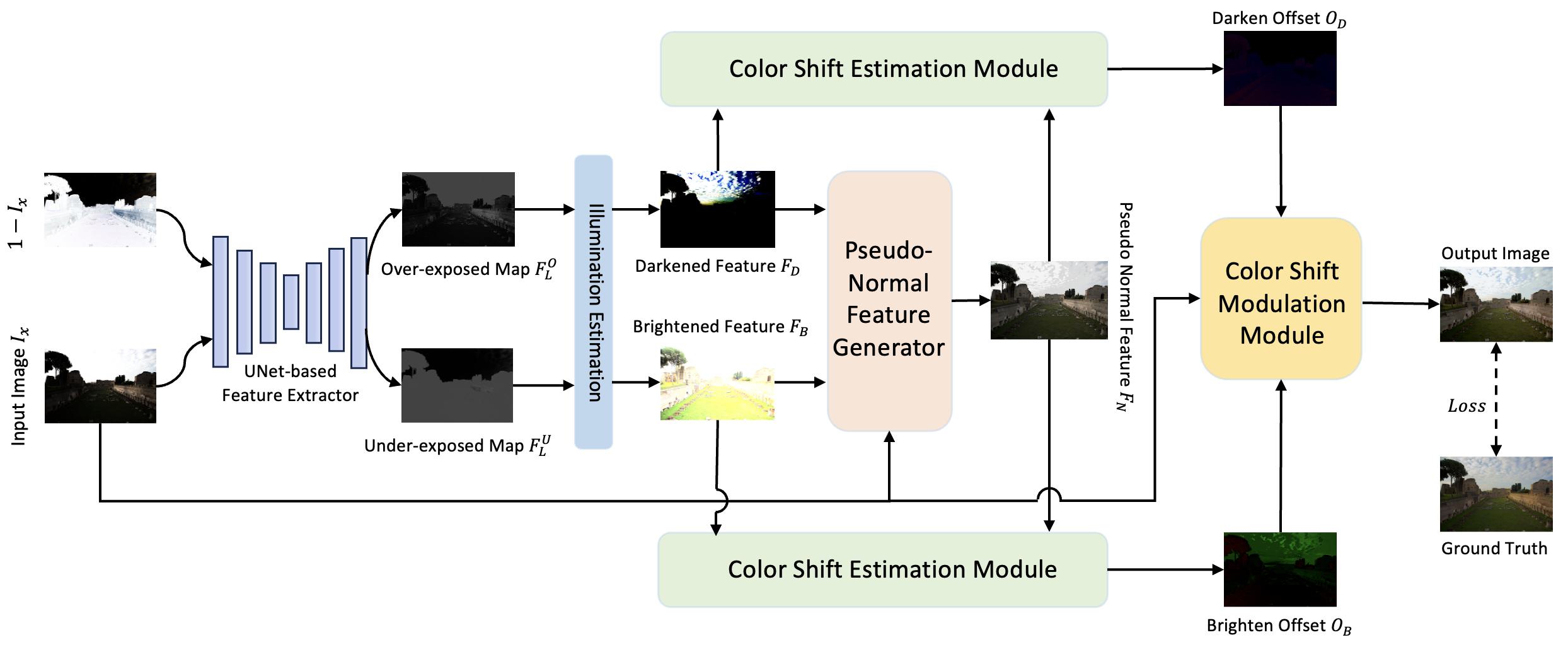}
    \vspace{-3mm}
    \caption{Overview of our proposed model. \ryn{We first generate} darkened features $F_D$ and brightened features $F_B$ using the UNet-based feature extractor. We then derive \ryn{a} pseudo-normal feature map $F_N$ using the generated brightened/darkened feature maps and the input image $I_x$. \ryn{We further} estimate the color shifts between the brightened/darkened color features $F_B$/$F_D$ and the created pseudo-normal feature map $F_N$ using the proposed Color Shift Estimation (COSE) module \ryn{to} obtain two individual offset maps $O_B$ and $O_D$. Finally, we modulate the image brightness and colors using the proposed Color Modulation (COMO) module, to produce the final output image.}
    \label{fig:pipeline}
    \vspace{-4mm}
\end{figure*}

\section{Proposed Method}
\label{methods}

Our method is inspired by two observations. First, the over-exposed pixels tend to have an inverse distribution shift compared to the under-exposed pixels. This suggests that it is necessary to capture and correct such color shifts separately. Second, \ryn{since~\kk{majority (if not all) of pixels are affected by over-/under-exposures}}, it is necessary to create pseudo-normal exposure information as guidance for the estimation of the color shifts of over-/under-exposed pixels.
Inspired by these two observations, we propose a novel network (Sec.~\ref{overview}) with two novel modules: a novel COlor Shift Estimation (COSE) module (Sec.~\ref{cose}) and a novel COlor MOdulation (COMO) module (Sec.~\ref{como}), for the enhancement of images with both over-/under-exposures.


\subsection{Network Overview}
\label{overview}
Given an input image $I_x\in \mathcal{R}^{3\times H\times W}$ with both over- and under-exposures, our network aims to produce an enhanced image $I_y\in \mathcal{R}^{3\times H\times W}$ with rectified image brightness and recovered image details and colors.
The overview of our model is shown in Figure~\ref{fig:pipeline}.
Given the input image $I_x$, we first compute its \kk{inverse}
version of $\hat{I}_x$ by $1-I_x$, both of which are fed into a UNet-based network to extract two illumination maps $F_L^U\in \mathcal{R}^{1\times H\times W}$ and $F_L^O\in \mathcal{R}^{1\times H\times W}$.
These two maps (\ie, $F_L^U$ and $F_L^O$) indicate the regions affected by under-exposure and over-exposure, respectively.
We then compute the darkened ($F_D$) and brightened ($F_B$) feature maps as:
\begin{align}
    F_B = \frac{I_x}{F_L^U} &= \frac{I_x}{f(I_x)}, \\
    F_D = 1-\frac{1-I_x}{F_L^O} &= 1 - \frac{1-I_x}{f(1 - I_x)},
\end{align}
where $f(\cdot )$ represents the UNet-based feature extractor. 
We model the color shifts based on the brightened and darkened feature maps $F_B, F_D\in \mathcal{R}^{3\times H\times W}$.

Given $F_B$ and $F_D$, we first apply a Pseudo-Normal Feature Generator to fuse them with the input image $I_x$ into a pseudo-normal feature map $F_N$, as:
\begin{align}
    F_N = g(F_B, F_D, I_x),
\end{align}
where $g(\cdot )$ denotes the pseudo \ryn{normal-exposure} generator. $F_N$ can then serve as a reference to guide the estimation of color shifts between $F_B$ and $F_N$ (and $F_D$ and $F_N$), separately, via two proposed COSE modules.
The darken offset $O_D$ and the brighten offset $O_B$ produced by the two COSE modules model both brightness and color shifts \ryn{w.r.t.} the input image $I_x$, and therefore $O_D$, $O_B$, and $I_x$ are fed into the proposed COMO module to adjust the image brightness and correct color shifts, to produce the final image $I_y$.


\subsection{Color Shift Estimation (COSE) Module}
\label{cose}

Unlike brightness adjustment, color shift correction is more challenging as it essentially requires the network to model the pixel directions in \yycr{RGB} color space, instead of the magnitudes of pixel intensity. Although there are some works~\cite{wang_2019_CVPR_deepupe, Xu_2021_TIP_deraining, Wang_2022_ECCV_lcdp} that use the cosine similarity regularization to help maintain the image colors during training, such a strategy often fails in large under-/over-exposed regions where pixels of small/high-intensity values are expected to have different colors.

\begin{figure}[b]
    \centering
    \includegraphics[width=\linewidth]{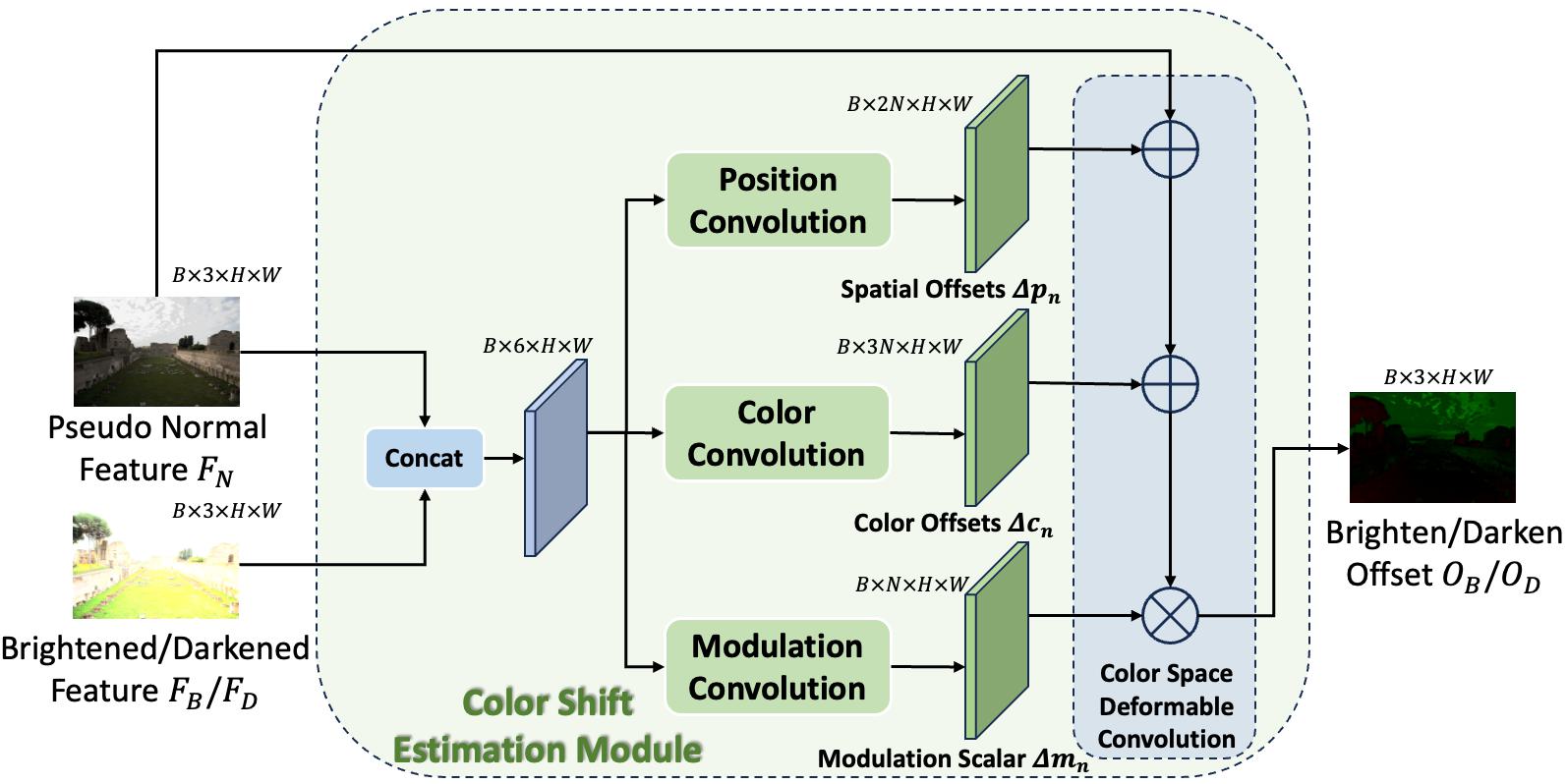}
    \caption{The proposed Color Shift Estimation (COSE) module aims to model the color distribution shifts in over- and under-exposed regions. We use three different convolutions to estimate the positional offsets, color shift offsets, and modulation scalars. We then utilize the color space deformable convolution in Eq.~\ref{eq:cdc} to compute the color offsets and produce the offset feature map for brightened and darkened features.
    }
    \label{fig:cose}
    \vspace{-4mm}
\end{figure}

We propose the COSE module to address this problem based on the deformable convolution techniques. \yycr{Deformable convolution (DConv) extends vanilla convolution by incorporating spatial offsets $\Delta p_n$ to adaptively perform convolution at any location of any $N\times N$ pixels, where $N\times N$ denotes the kernel size. The modulation $\Delta m_n$ is proposed to assign different weights to different kernel positions, making the convolution operator focusing on important pixels.} While DConv can predict the offsets with respect to the \rynq{basis}, it is possible for DConv to capture the color distribution shift. However, since previous methods~\cite{dai_2017_deformable, zhu_2019_deformablev2, Wang_2023_deformablev3} only apply DConv in the pixel spatial domain, we propose to extend DConv to be performed in both \rynq{spatial domain} and color space, for modeling the brightness changes and color shifts jointly.
%
Specifically, as depicted in Figure~\ref{fig:cose}, our COSE module first concatenates the pseudo normal feature map $F_N$ and the brightened/darkened feature map $F_B$/$F_D$ along the channel dimension. 
It then employs three separate $3\times 3$ convolutions to extract positional offsets $\Delta p_n\in \mathcal{R}^{B\times 2N\times H\times W}$, color offsets $\Delta c_n\in \mathcal{R}^{B\times 3N\times H\times W}$, and modulation $\Delta m_n\in \mathcal{R}^{B\times N\times H\times W}$. 
Positional offsets $\Delta p_n$ and modulation $\Delta m_n$ are the same as those in~\cite{zhu_2019_deformablev2}. \ryn{They are} performed in the \rynq{spatial domain } to aggregate spatial contextual information within the deformed irregular receptive fields of the convolution operation.
Additionally, the color offsets $\Delta c_n$ are introduced, to represent the color offsets for each channel at each kernel position.
The learned color offsets $\Delta c_n$ are formulated to have $3N$ channels to model the color shifts of the input sRGB image with 3 channels.

Formally, the computation of deformable convolution in both spatial domain and color space can be written as:
\begin{align}
    y = \sum_{p_n\in \mathcal{R}} (w_n\cdot x(p_0 + p_n + \Delta p_n) + \Delta c_n) \cdot \Delta m_n,\label{eq:cdc}
\end{align}
where \yycr{$x$ denotes the input features to the convolution operation, while $p_0$, $p_n$, and $\Delta p_n$ are 2-dimensional variables indicating the spatial location, and} $y$ (or $y(p_0)$) denotes the output of color space deformable convolution for each pixel $p_0$ within the input image. The set $\mathcal{R} = \{(-1, -1), (-1, 0), \dots, (1, 1)\}$ represents the grids of a regular $3\times 3$ kernel. $n$ is the enumerator for the elements in $\mathcal{R}$ indicating the $n$-th position, and $N$ is the length of $\mathcal{R}$ ($N=9$ for a regular $3\times 3$ kernel).
Since the displacement $\Delta p_n$ may assume fractional values in practice, we apply bilinear interpolation for computation, which aligns with the spatial deformable convolution~\cite{dai_2017_deformable}.


\subsection{Color Modulation (COMO) Module}
\label{como}

We propose the COMO module to modulate the brightness and color of the input image and produce the final output image $I_y$, based on the learned offsets $O_B$/$O_D$ between brightened features $F_B$/darkened features $F_D$ and pseudo normal features $F_N$. Since it is crucial to aggregate global information in order to produce the corrected image with harmonious colors, we draw inspiration from the non-local context modeling in~\cite{zhu_2021_nonlocal} and formulate our COMO module by extending the self-affinity computation in~\cite{zhu_2021_nonlocal} into the cross-affinity computation so that COMO can enhance the input image by querying both $O_B$ and $O_D$.

\begin{figure}[t]
    \centering
    \includegraphics[width=\linewidth]{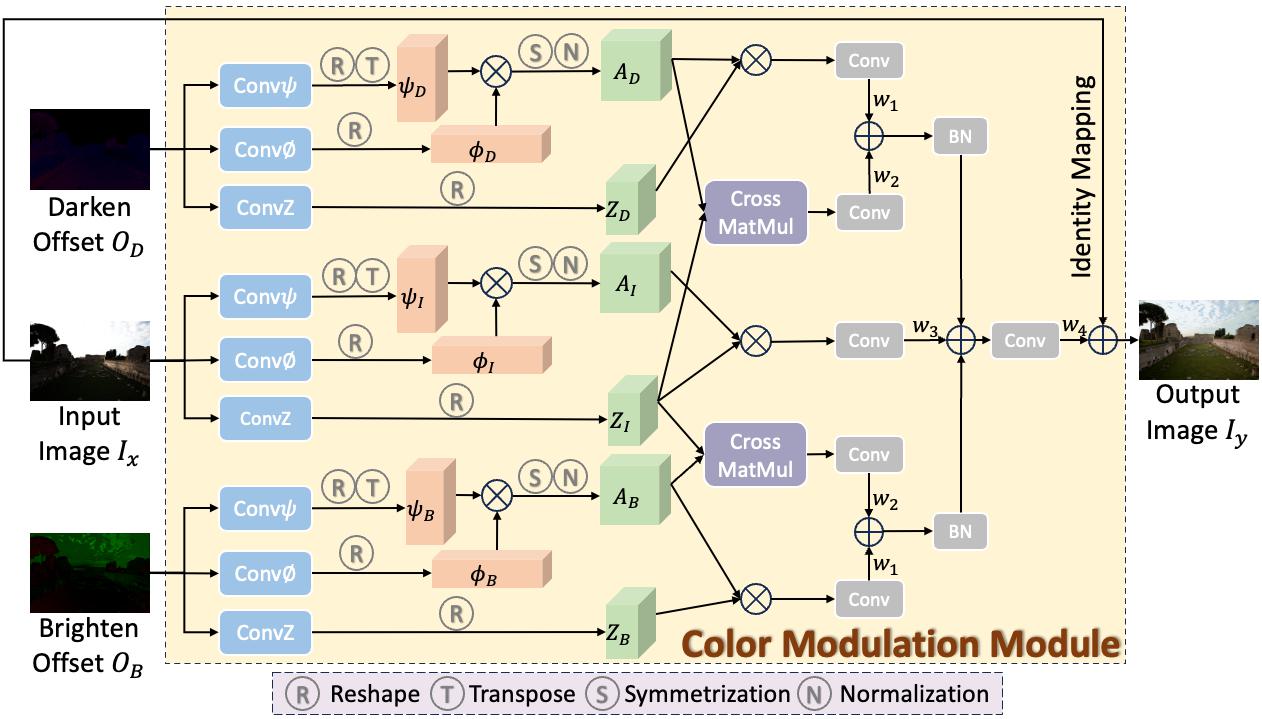}
    \vspace{-3mm}
    \caption{The proposed Color Modulation (COMO) module aims to adjust the brightness and correct the color shifts of the input image $I_x$, guided by the color offsets \ryn{from} our COSE modules.
    The COMO module takes the image $I_x$ and brighten/darken offsets $O_B$/$O_D$ as input and produces the enhanced image via a tailored cross-attention mechanism.
    }
    \label{fig:como}
    \vspace{-5mm}
\end{figure}

As shown in Figure~\ref{fig:como}, we assign three branches to process the input image $I_x$, darken offset $O_D$, and brighten offset $O_B$, separately, while each branch contains three $1\times 1$ convolution layers (denoted as $Conv\psi$, $Conv\phi$, and $ConvZ$).
We then compute the self-affinity matrix $A_i$ in each branch, as:
\begin{align}
    A_i = \psi_i \otimes \phi_i,\ for\ i\in \{I, B, D\},
\end{align}
where $\otimes$ is the matrix multiplication, $\psi_i$ and $\phi_i$ are the feature maps obtained by $Conv\psi$ and $Conv\phi$, respectively. $A_i$ is then symmetrized and normalized to ensure the existence of real eigenvalues and the stabilization of backward propagation. Each row of $A_i$ serves as a spatial attention map, and $Z_i$ (obtained by $ConvZ$) serves as weights for the attention maps.
Next, we model the correlations between $I_x$ and $O_B$/$O_D$ via matrix multiplication and add them together with the \yy{self-affined} features, as:
\begin{align}\label{eq:cross}
    f_j = w_1 A_j \otimes Z_j + w_2 A_j \otimes Z_I,
\end{align}
where $j\in \{B, D\}$ is the index of the affinity matrix $A_j$ and feature map $Z_j$ from the brighten or darkened branch. $w_1$ and $w_2$ are weight matrices generated by $1\times 1$ convolutions.
While the first term in Eq.~\ref{eq:cross} is to~\yy{discover the significant color offset regions in $O_B$ and $O_D$ learned by COSE, and} the second term aims to~\yy{use the learned weights of input $Z_I$ to attend to attention maps of $O_B$ and $O_D$ to discover what the offsets are like in the significant regions of input.}

Finally, we combine $f_B$, $f_D$ and input image $I_x$ \yy{to fuse the explored color offsets as guidance for input image.} We then produce the final result $I_y$, as:
\begin{align}
    I_y = w_4(BN(f_B) + BN(f_D) + w_3A_I\otimes Z_I) + I_x,
\end{align}
where $BN(\cdot )$ represents Batch Normalization, and $w_3$, $w_4$ are weight matrices generated by $1\times 1$ convolutions.

\subsection{Loss Function}

Two loss functions, $\mathcal{L}_{pesudo}$ and $\mathcal{L}_{output}$ are used to train our network. As we need to generate a pseudo-normal feature map to help identify the color shifts, we use $\mathcal{L}_{pesudo}$ to provide intermediate supervision for the generation process\yycr{:}
\begin{align}
    \mathcal{L}_{pesudo} = ||F_N - GT||_1.
\end{align}

The $\mathcal{L}_{output}$ contains four terms to supervise the network to produce the enhanced images, which are $L1$ loss, Cosine similarity $\mathcal{L}_{cos}$, SSIM loss $\mathcal{L}_{ssim}$~\cite{wang_2004_TIP_SSIM}, and VGG loss $\mathcal{L}_{vgg}$~\cite{simonyan_2014_vgg}.
The $\mathcal{L}_{output}$ can be formulated as:
\begin{align}\label{eq:loss1}
    \mathcal{L}_{output} = \lambda_1 \mathcal{L}_{L1} + \lambda_2 \mathcal{L}_{cos} + \lambda_3 \mathcal{L}_{ssim} + \lambda_4 \mathcal{L}_{vgg},
\end{align}
where $\lambda_1$, $\lambda_2$, $\lambda_3$ and $\lambda_4$ are four balancing hyper-parameters. The overall loss function is then:
\begin{align}\label{eq:loss2}
    \mathcal{L} = \lambda_p \mathcal{L}_{pesudo} + \lambda_o \mathcal{L}_{output},
\end{align}
where $\lambda_p$ and $\lambda_o$ are two balancing hyper-parameters.

\label{loss}

%% file: sec/4_experiments.tex
\section{Experiments}
\label{exp}

\subsection{Experimental Settings}

\noindent \textbf{Datasets.}\quad {We evaluate our proposed method on two datasets: LCDP~\cite{Wang_2022_ECCV_lcdp}, and MSEC~\cite{Afifi_2021_CVPR_MSEC}}, both of which are derived from the MIT-Adobe FiveK dataset~\cite{bychkovsky_2011_CVPR_Adobe5K}. Specifically, LCDP contains images with both over-exposure and under-exposure regions, and we use it to test the ability of our method to handle both over- and under-exposure conditions. The LCDP dataset consists of 1,733 images, which are split into 1,415 for training, 100 for validation, and 218 for testing. MSEC contains images with either over- or under-exposure, while their exposure values are in the collection of \{-1.5EV, -1EV, 0EV, +1EV, +1.5EV\}, and the expert-retouched images are used as the ground truths. The MSEC dataset comprises 17,675 training images, 750 validation images, and 5,905 testing images. We conduct experiments on this dataset in order to test the generalization ability of the proposed method.

\noindent \textbf{Evaluation Metrics.}\quad To evaluate the performance of different image enhancement methods and our proposed method, we adopt the commonly used metrics Peak Signal-to-Noise Ratio (PSNR) and Structural Similarity Index Measure (SSIM)~\cite{wang_2004_TIP_SSIM} as numerical evaluation metrics. \yycr{Additionally, to gauge the accuracy of color correction, we utilize the Root Mean Squared Error (RMSE) on LAB color space as a color-related metric.}

\subsection{Comparisons with State-of-the-art Methods}

\noindent{\bf Quantitative Comparisons.} To verify the effectiveness of our proposed method, we compare our method with one conventional method: Histogram Equalization~\cite{pizer_1987_histogram_equalization}, and \kk{13} deep learning based SOTA methods: DSLR~\cite{ignatov_2017_dslr}, HDRNet~\cite{gharbi_2017_hdrnet}, RetinexNet~\cite{wei_2018_BMVC_retinexnet}, DeepUPE~\cite{wang_2019_CVPR_deepupe}, ZeroDCE~\cite{Guo_2020_CVPR_Zero-DCE}, MSEC~\cite{Afifi_2021_CVPR_MSEC}, RUAS~\cite{liu_2021_CVPR_ruas}, SNRNet~\cite{Xu_2022_CVPR_SNR}, 
LCDPNet~\cite{Wang_2022_ECCV_lcdp}, FecNet~\cite{Huang_2022_ECCV_Fourier}, LANet~\cite{yang_2023_lanet}, SMG~\cite{Xu_2023_CVPR_Structure}, and RetinexFormer~\cite{cai_2023_retinexformer}.

\begin{table}[b]
    \centering
    \resizebox{0.7\columnwidth}{!}{
    \begin{tabular}{l|cc|c}
    \hhline{====}
       Method  & PSNR~$\uparrow$ & SSIM~$\uparrow$ & \# params \\
       \hline
       HE'87~\cite{pizer_1987_histogram_equalization} & 16.215 & 0.669 & - \\
       DSLR'17~\cite{ignatov_2017_dslr} & 20.856 & 0.758 & 0.39M \\
       HDRNet'17~\cite{gharbi_2017_hdrnet} & 21.834 & 0.818 & 0.48M \\
       RetinexNet'18~\cite{wei_2018_BMVC_retinexnet} & 20.199 & 0.709 & 0.84M \\
       DeepUPE'19~\cite{wang_2019_CVPR_deepupe} & 20.970 & 0.818 & 1.02M \\
       ZeroDCE'20~\cite{Guo_2020_CVPR_Zero-DCE} & 12.861 & 0.668 & {0.08M} \\
       MSEC'21~\cite{Afifi_2021_CVPR_MSEC} & 20.377 & 0.779 & 7.04M \\
       RUAS'21~\cite{liu_2021_CVPR_ruas} & 13.757 & 0.606 & {0.003M} \\
       SNRNet'22~\cite{Xu_2022_CVPR_SNR} & 20.829 & 0.711 & 4.01M \\
       LCDPNet'22~\cite{Wang_2022_ECCV_lcdp} & 22.931 & 0.817 & 0.28M \\
       FECNet'22~\cite{Huang_2022_ECCV_Fourier} & 23.333 & 0.823 & 0.15M \\
       LANet'23~\cite{yang_2023_lanet} & 21.444 & 0.691 & 0.57M \\
       \kk{SMG'23}~\cite{Xu_2023_CVPR_Structure} & 22.427 & 0.786 & 17.90M \\
       RetinexFormer'23~\cite{cai_2023_retinexformer} & \underline{23.360} & \underline{0.850} & 1.61M \\
       \hline
       Ours & \textbf{23.627} & \textbf{0.855} & 0.30M \\
    \hhline{====}
    \end{tabular}}
    \caption{Quantitative comparison between the proposed method and SOTA methods on the LCDP~\cite{Wang_2022_ECCV_lcdp} test set. All methods are re-trained on the LCDP training set. Best results are marked in \textbf{bold} and second best results are \underline{underlined}.}
    \label{tab:quantitative-comparison-lcdp}
    \vspace{-3mm}
\end{table}


We first evaluate the effectiveness of our model in enhancing images with both over- and under-exposures by comparing our model to existing methods on the LCDP~\cite{Wang_2022_ECCV_lcdp} dataset. Table~\ref{tab:quantitative-comparison-lcdp} reports the comparison results, where all methods are re-trained for a fair comparison.
We can see that our method outperforms the second-best method, the RetinexFormer~\cite{cai_2023_retinexformer} in terms of the PSNR and SSIM metrics while using only around $20\%$ network parameters compared to it.
Our method outperforms the LCDPNet~\cite{Wang_2022_ECCV_lcdp} by a large margin ($\Delta$PSNR: $+0.70$, $\Delta$SSIM: $+0.038$) while we have approximately similar network parameters (Ours/LCDPNet: 0.30M/0.28M).

\begin{table}[t]
    \centering
    \resizebox{0.7\columnwidth}{!}{
    \begin{tabular}{l|cccc}
    \hhline{=====}
       Method & HE~\cite{pizer_1987_histogram_equalization} & Retinex~\cite{wei_2018_BMVC_retinexnet} & ZeroDCE~\cite{Guo_2020_CVPR_Zero-DCE} \\
       RMSE~$\downarrow$ & 7.525 & 6.945 & 7.960 \\
       \hline
       Method & MSEC~\cite{Afifi_2021_CVPR_MSEC} & SNR~\cite{Xu_2022_CVPR_SNR} & LCDP~\cite{Wang_2022_ECCV_lcdp} \\
       RMSE~$\downarrow$ & 7.378 & 6.715 & 6.608 \\
       \hline
       Method & FEC~\cite{Huang_2022_ECCV_Fourier} & RetinexF~\cite{cai_2023_retinexformer} & Ours \\
       RMSE~$\downarrow$ & 6.384 & \underline{6.148} & \textbf{6.105} \\
    \hhline{=====}
    \end{tabular}}
    \caption{\yycr{Quantitative comparison between the proposed method and SOTA methods using RMSE on the LCDP~\cite{Wang_2022_ECCV_lcdp} test set. All methods are re-trained on the LCDP training set. Best performances are marked in \textbf{bold} and second best results are \underline{underlined}.}}
    \label{tab:quantitative-comparison-rmse}
    \vspace{-3mm}
\end{table}

\yycr{To evaluate the color correction performance of our method, we also compare our model with SOTA methods using color-related metrics, which is the root mean squared error (RMSE) on LAB color space. Table~\ref{tab:quantitative-comparison-rmse} shows the results, where our method achieves the best performance.}

\begin{figure*}[t]
\renewcommand{\tabcolsep}{0.8pt}
\begin{center}

\begin{tabular}{ccccc}

    \includegraphics[width=0.19\linewidth]{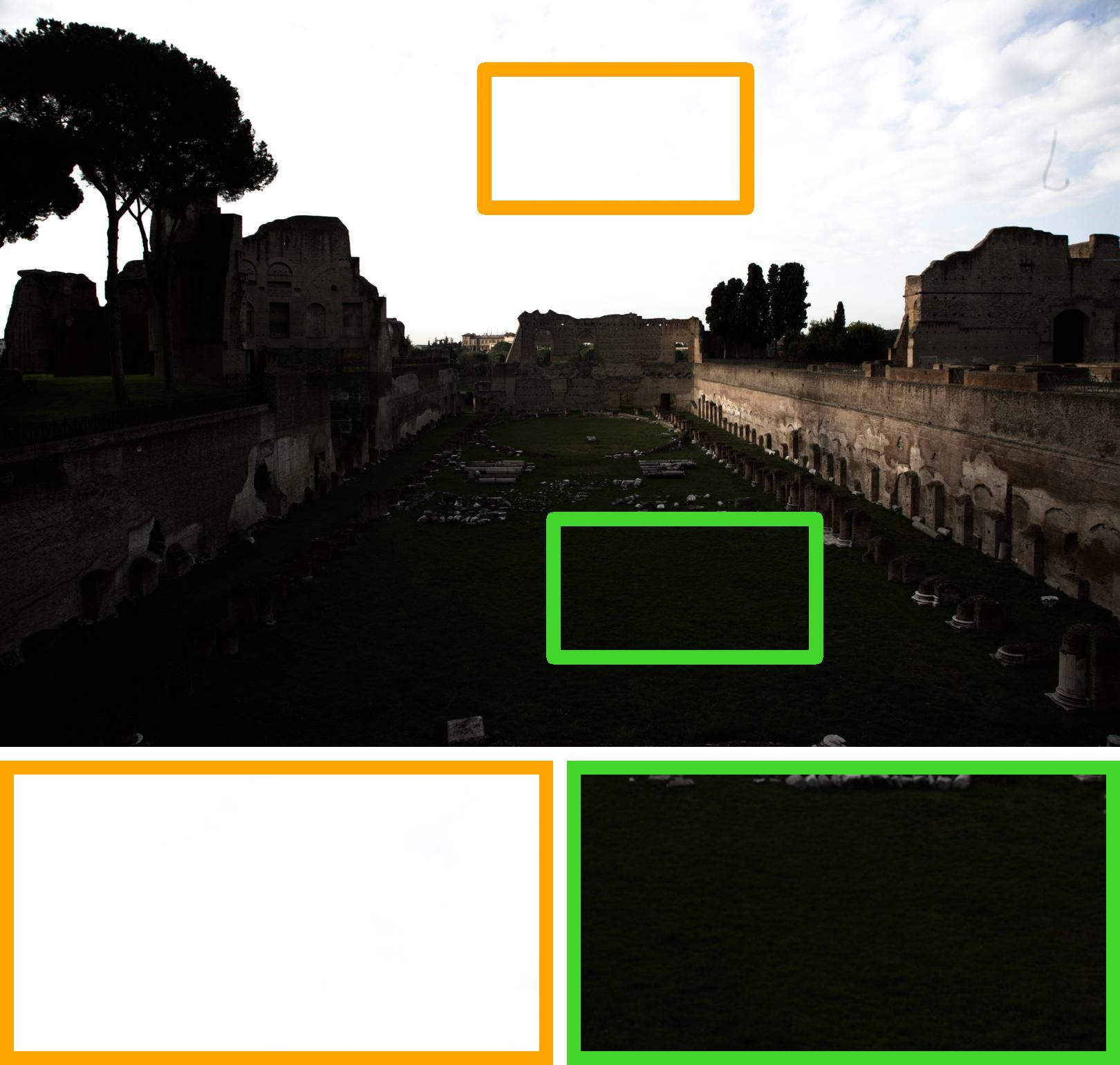} &
    \includegraphics[width=0.19\linewidth]{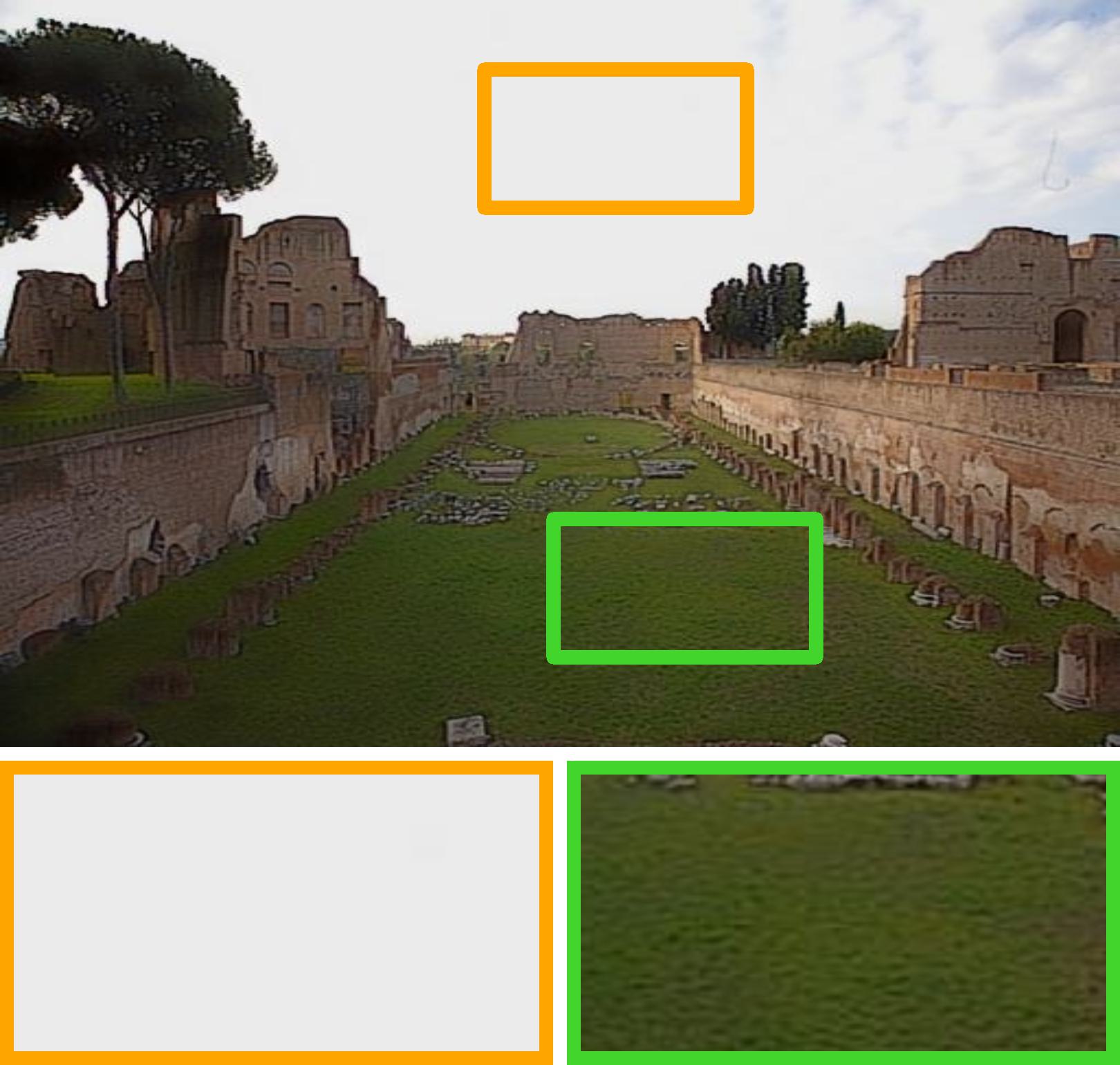} &
    \includegraphics[width=0.19\linewidth]{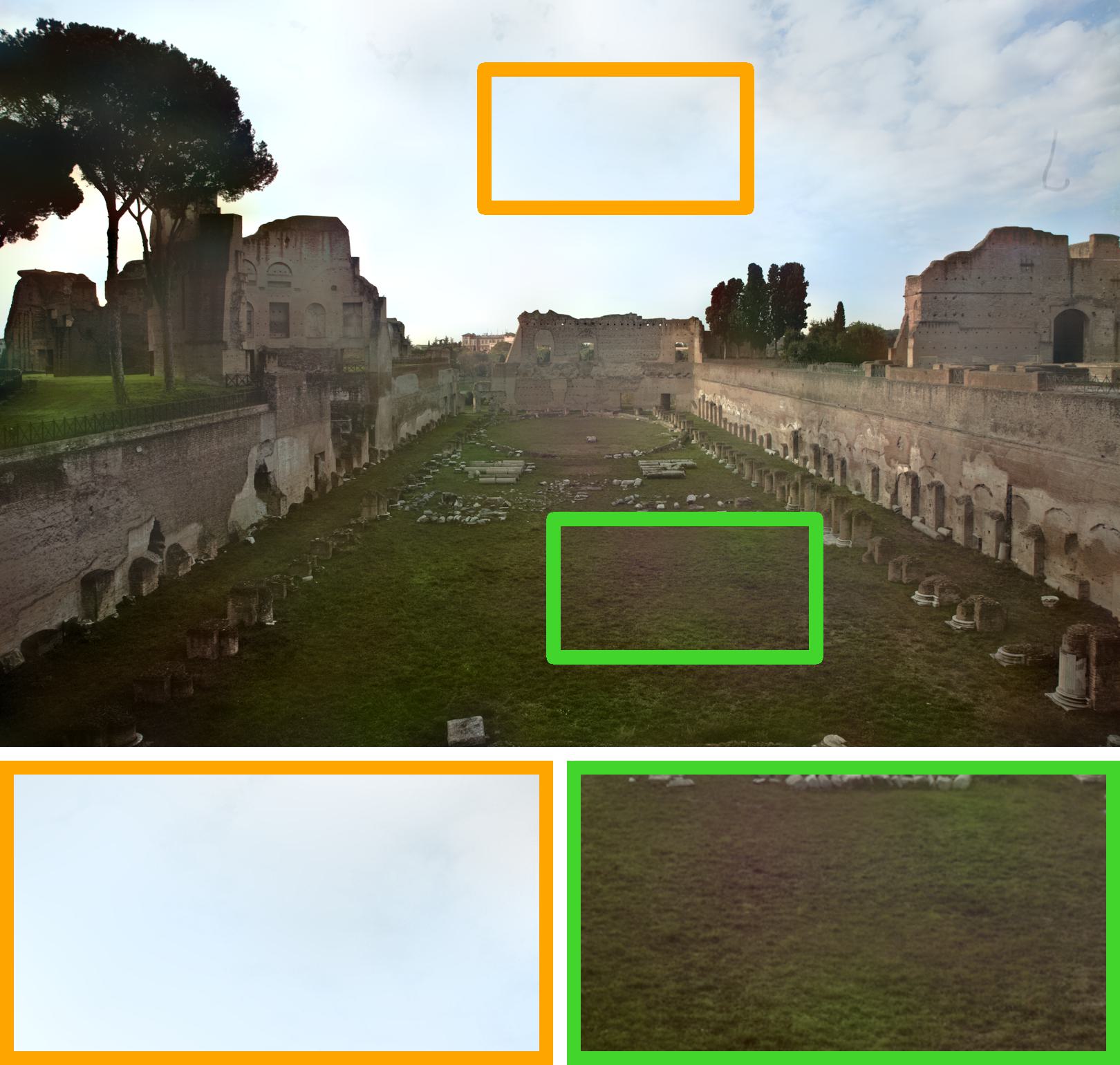} &
    \includegraphics[width=0.19\linewidth]{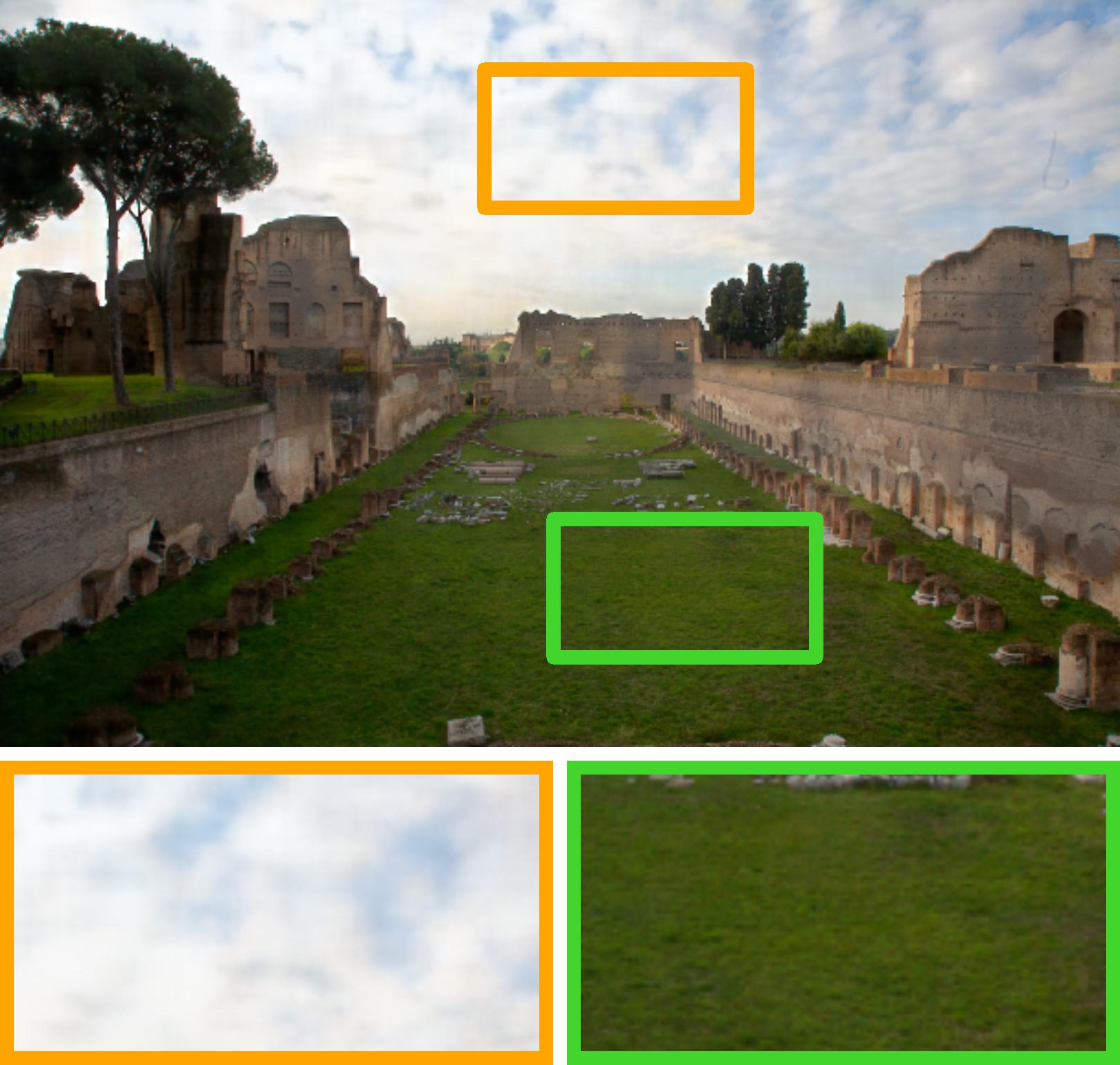} &
    \includegraphics[width=0.19\linewidth]{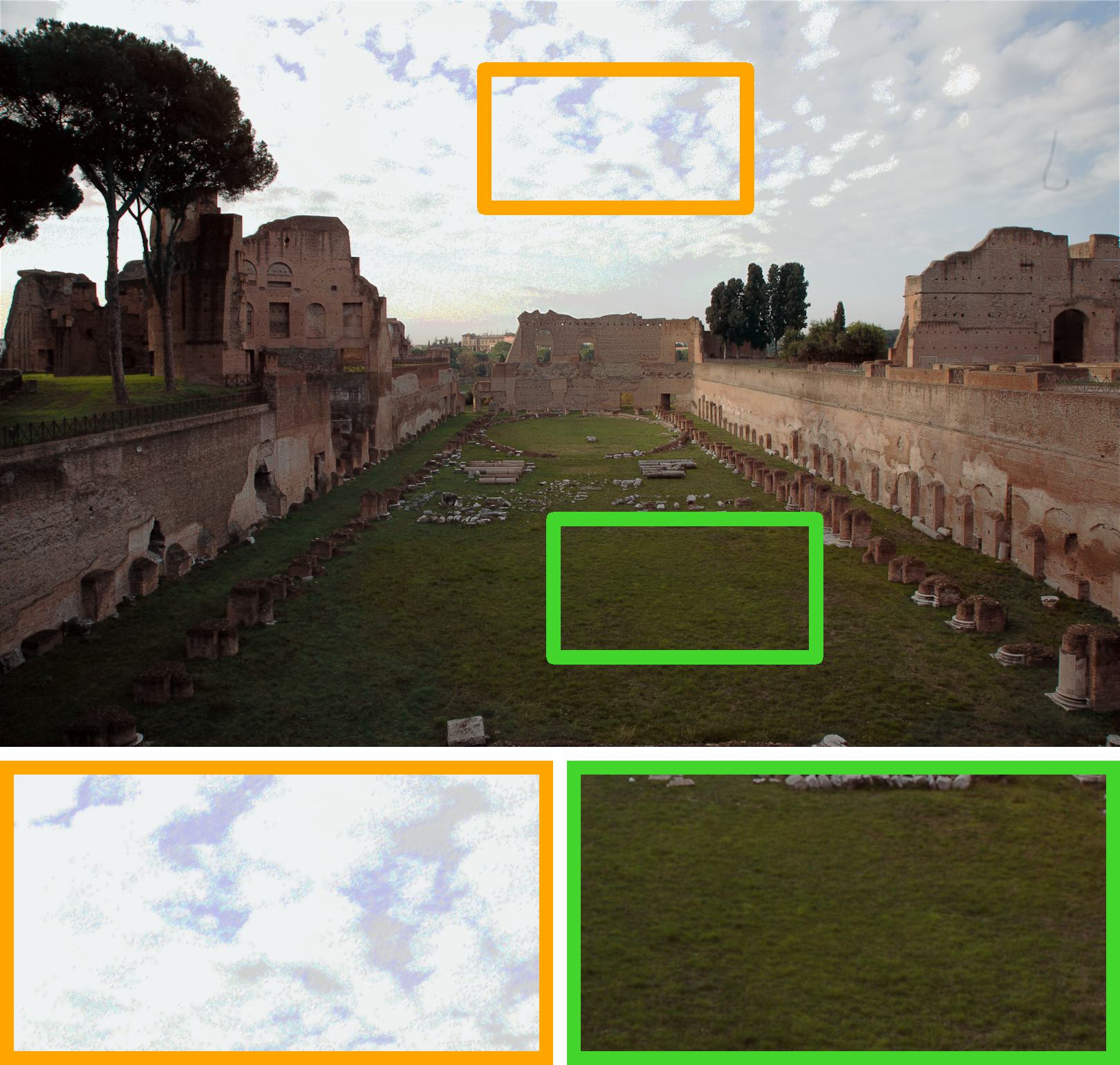}\\ 
    \scriptsize{(a) Input} & \scriptsize{(b) RetinexNet~\cite{wei_2018_BMVC_retinexnet}}
    &\scriptsize{(c) MSEC~\cite{Afifi_2021_CVPR_MSEC}} & \scriptsize{(d) SNRNet~\cite{Xu_2022_CVPR_SNR}} & \scriptsize{(e) LCDP~\cite{Wang_2022_ECCV_lcdp}}\\

    \includegraphics[width=0.19\linewidth]{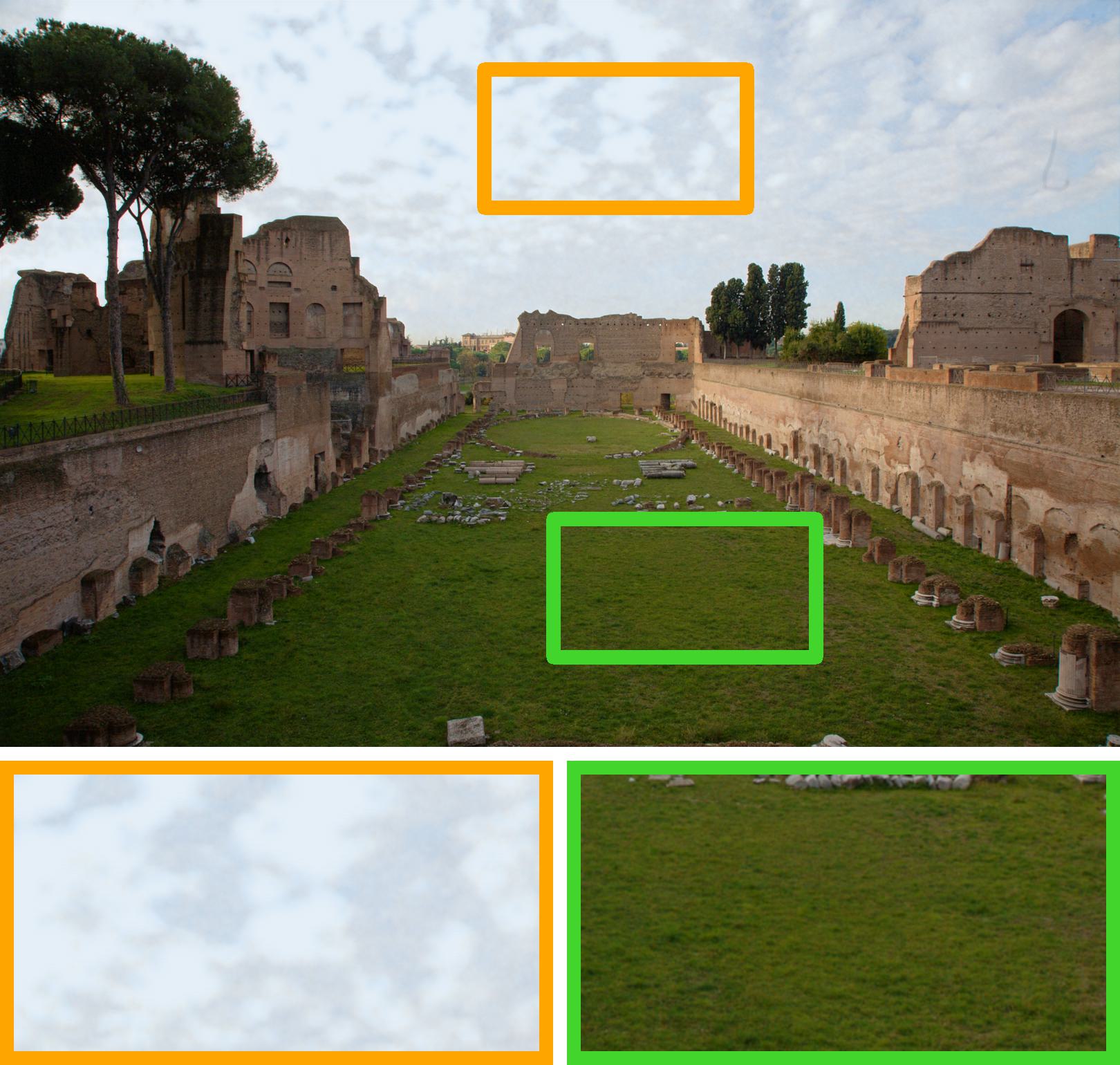} &
    \includegraphics[width=0.19\linewidth]{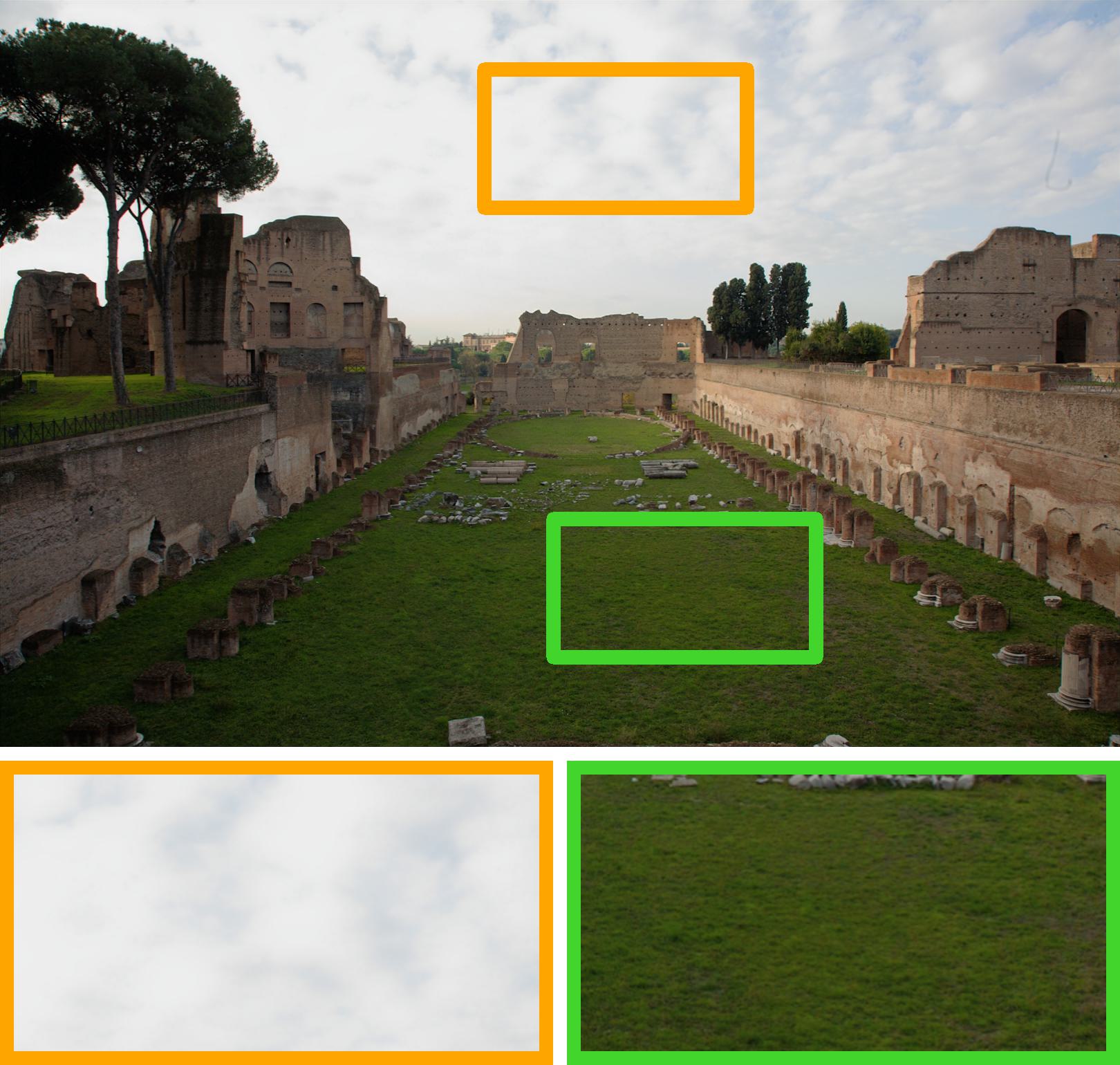} &
    \includegraphics[width=0.19\linewidth]{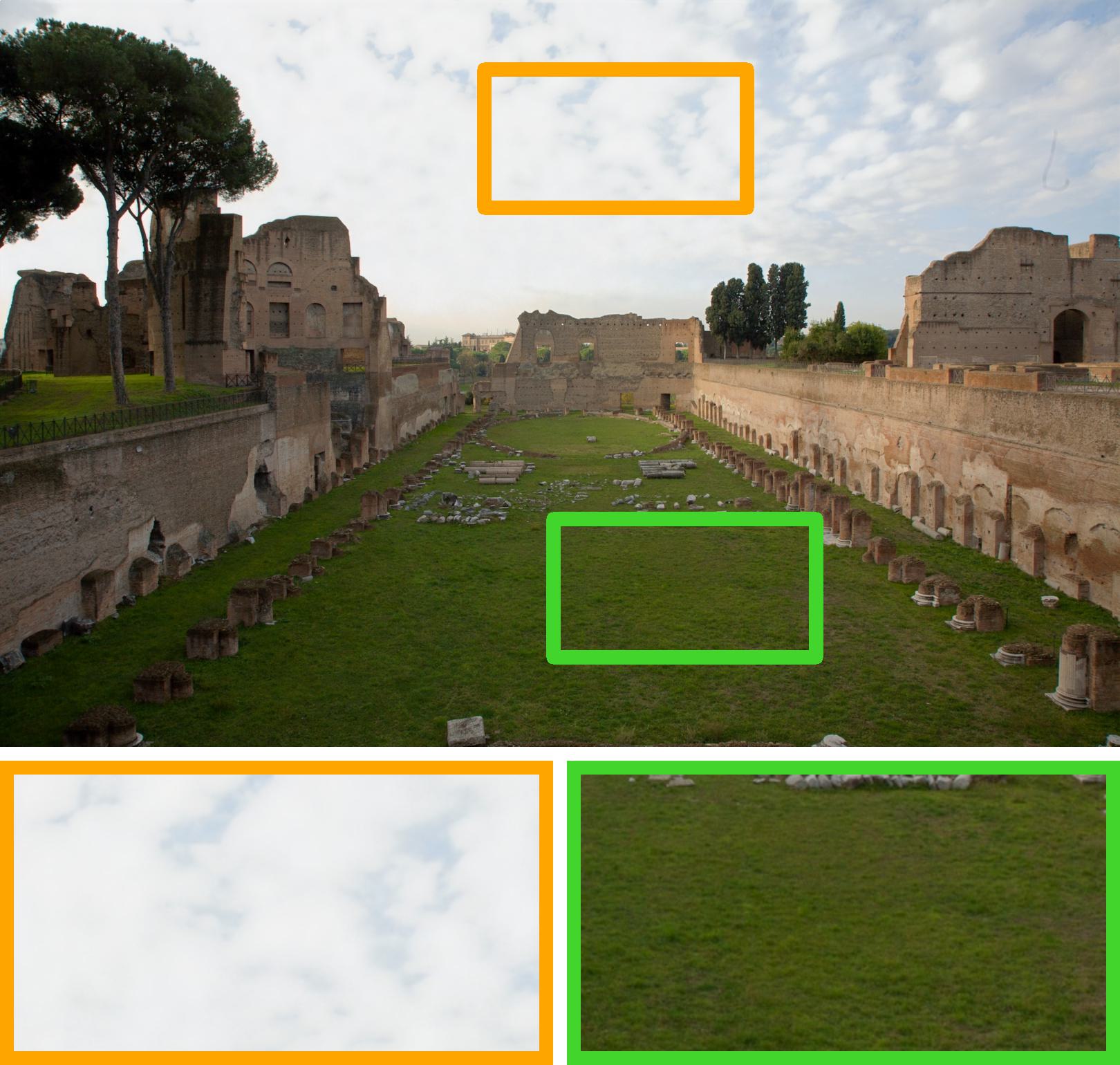} &
    \includegraphics[width=0.19\linewidth]{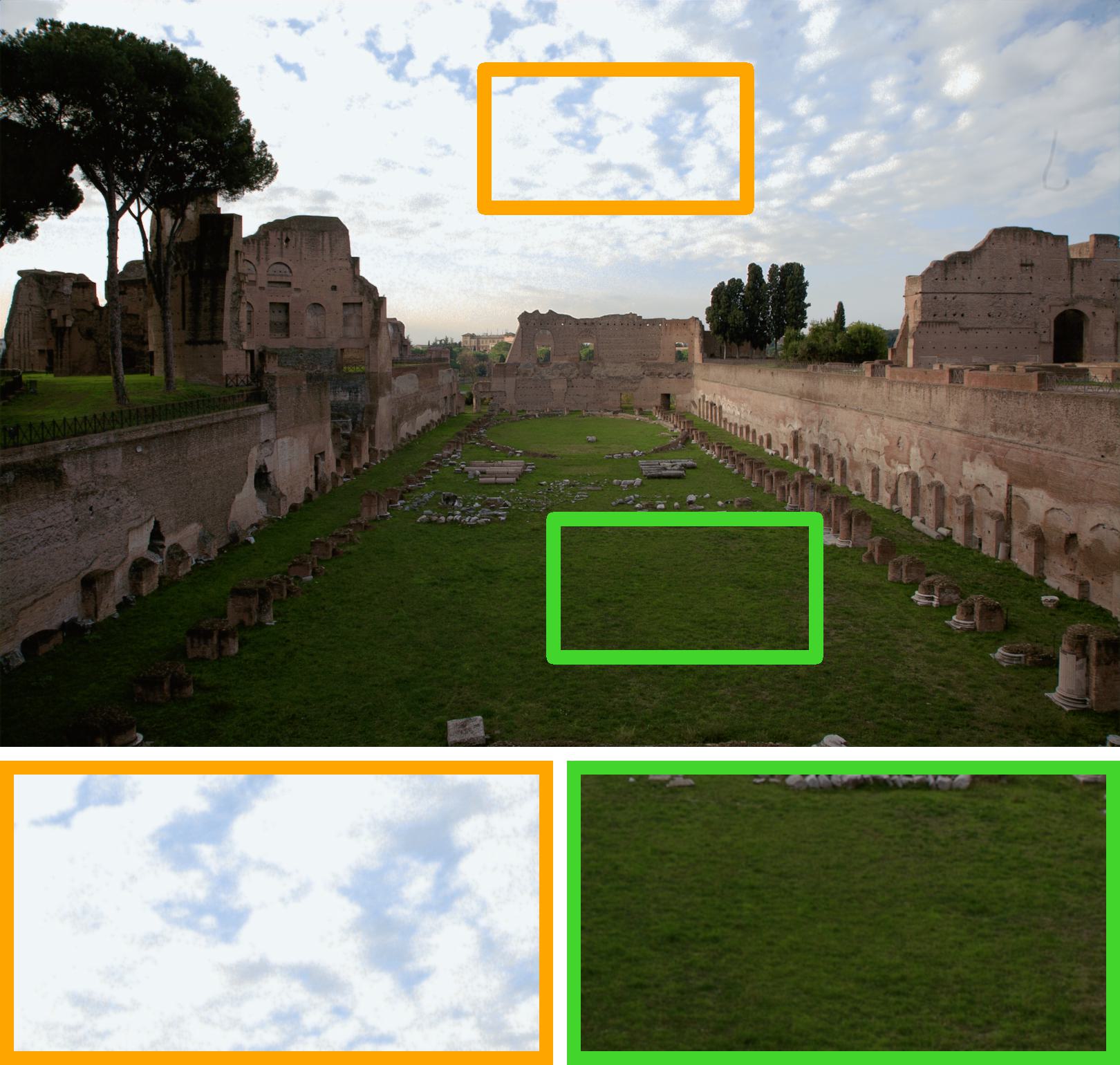} &
    \includegraphics[width=0.19\linewidth]{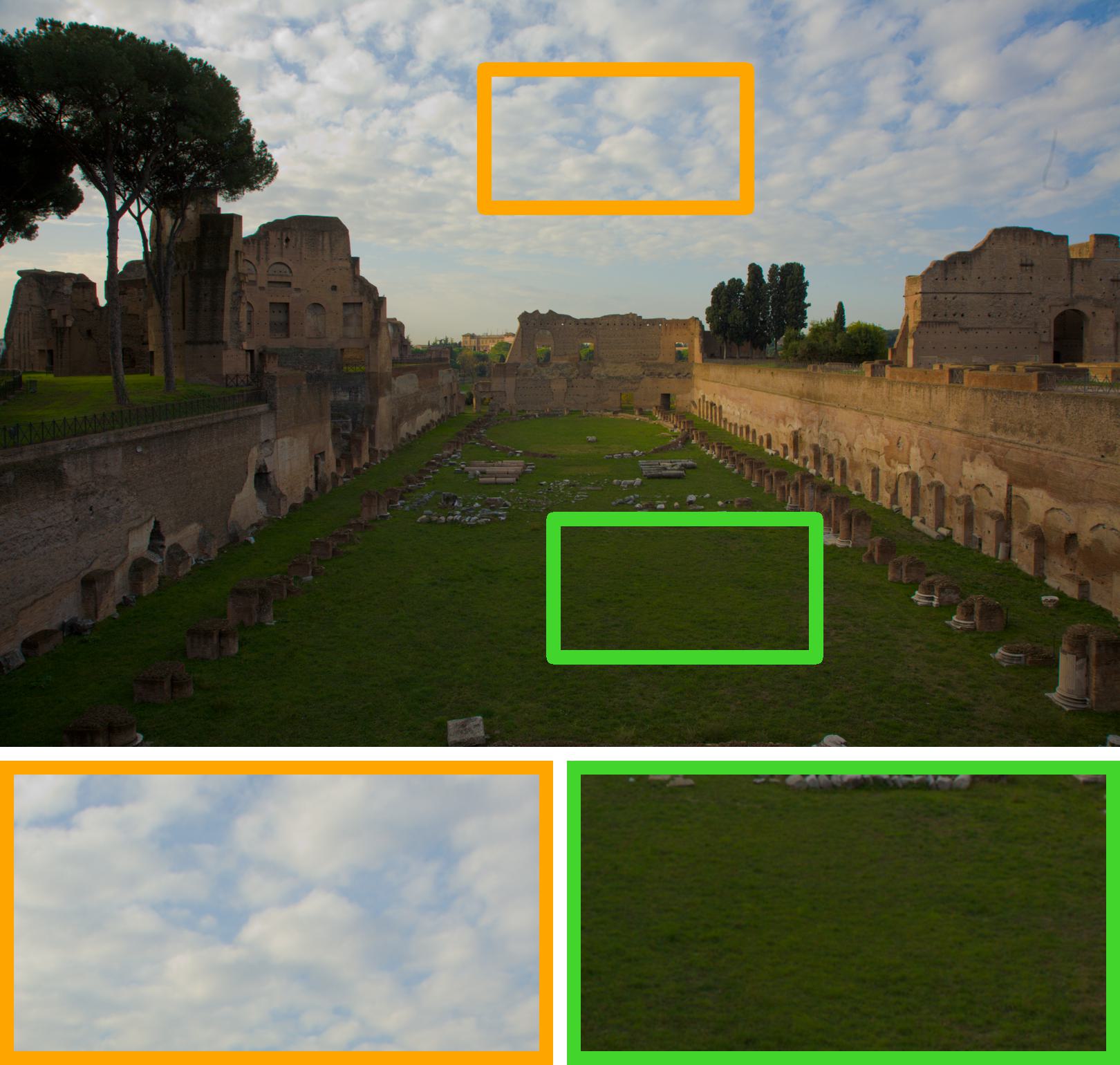}\\ 
    \scriptsize{(f) RetinexFormer~\cite{cai_2023_retinexformer}} & \scriptsize{(g) LANet~\cite{yang_2023_lanet}}
    &\scriptsize{(h) FECNet~\cite{Huang_2022_ECCV_Fourier}} & \scriptsize{(i) Ours} & \scriptsize{(j) GT}\\

    \includegraphics[width=0.19\linewidth]{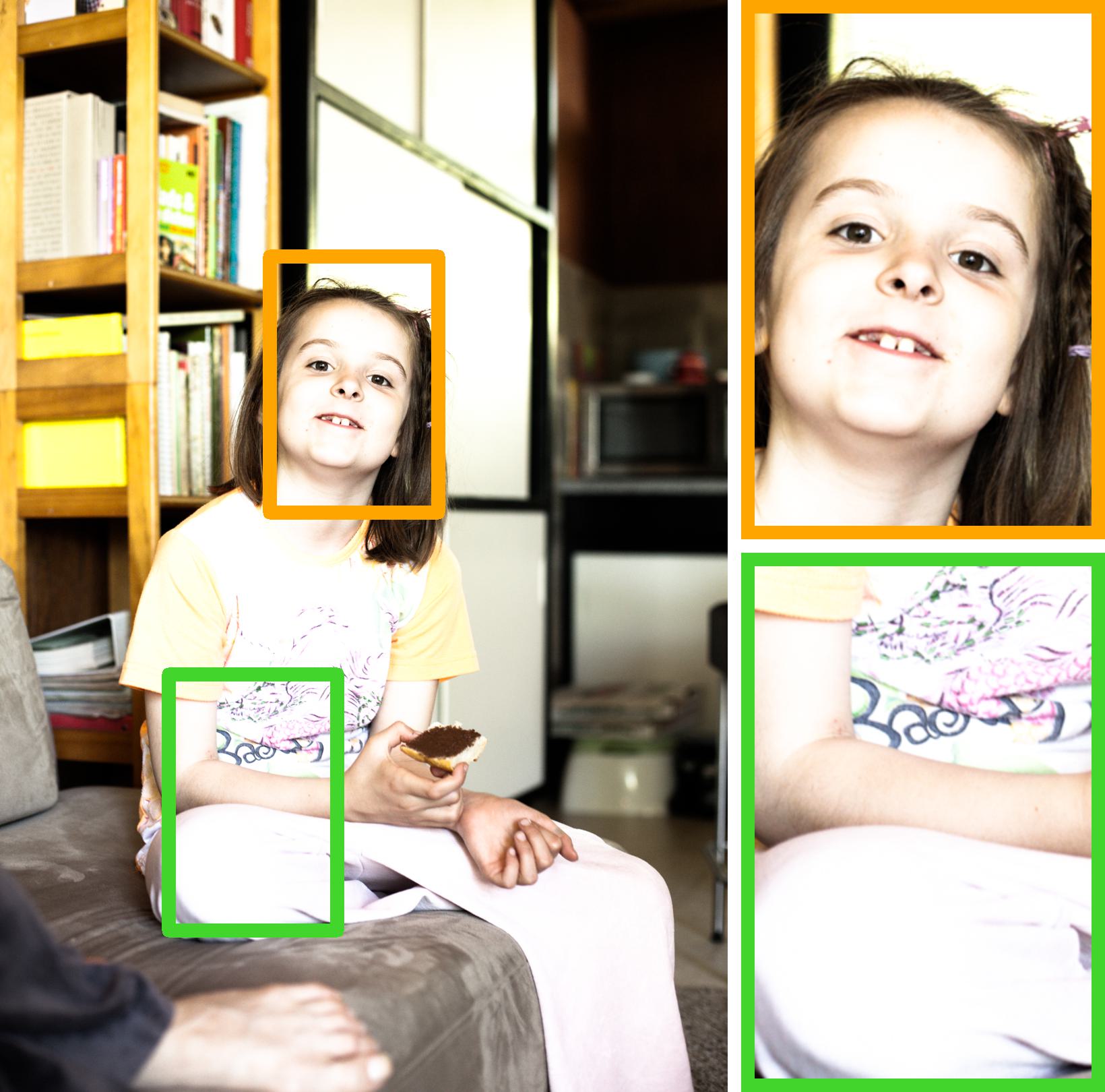} &
    \includegraphics[width=0.19\linewidth]{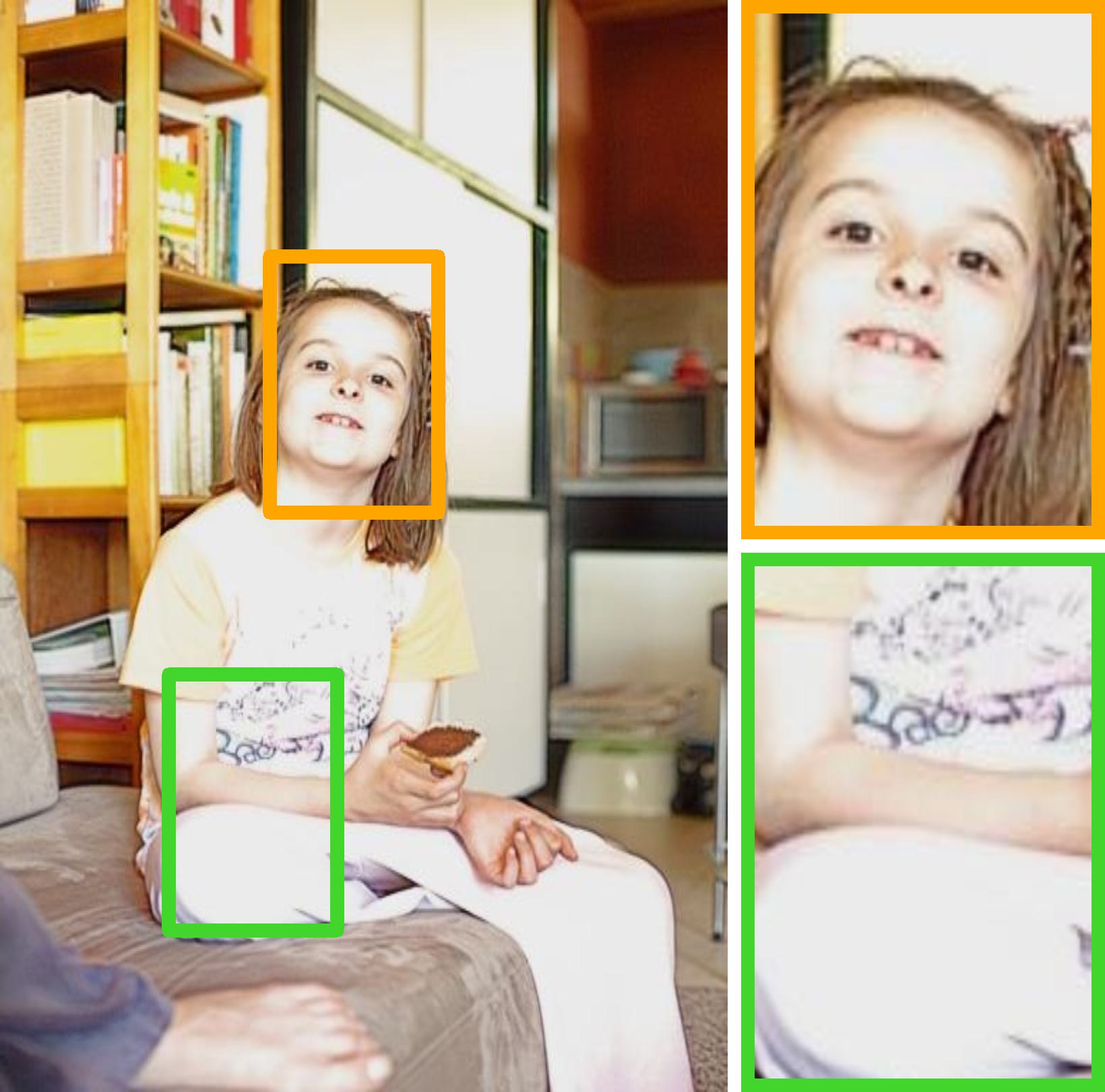} &
    \includegraphics[width=0.19\linewidth]{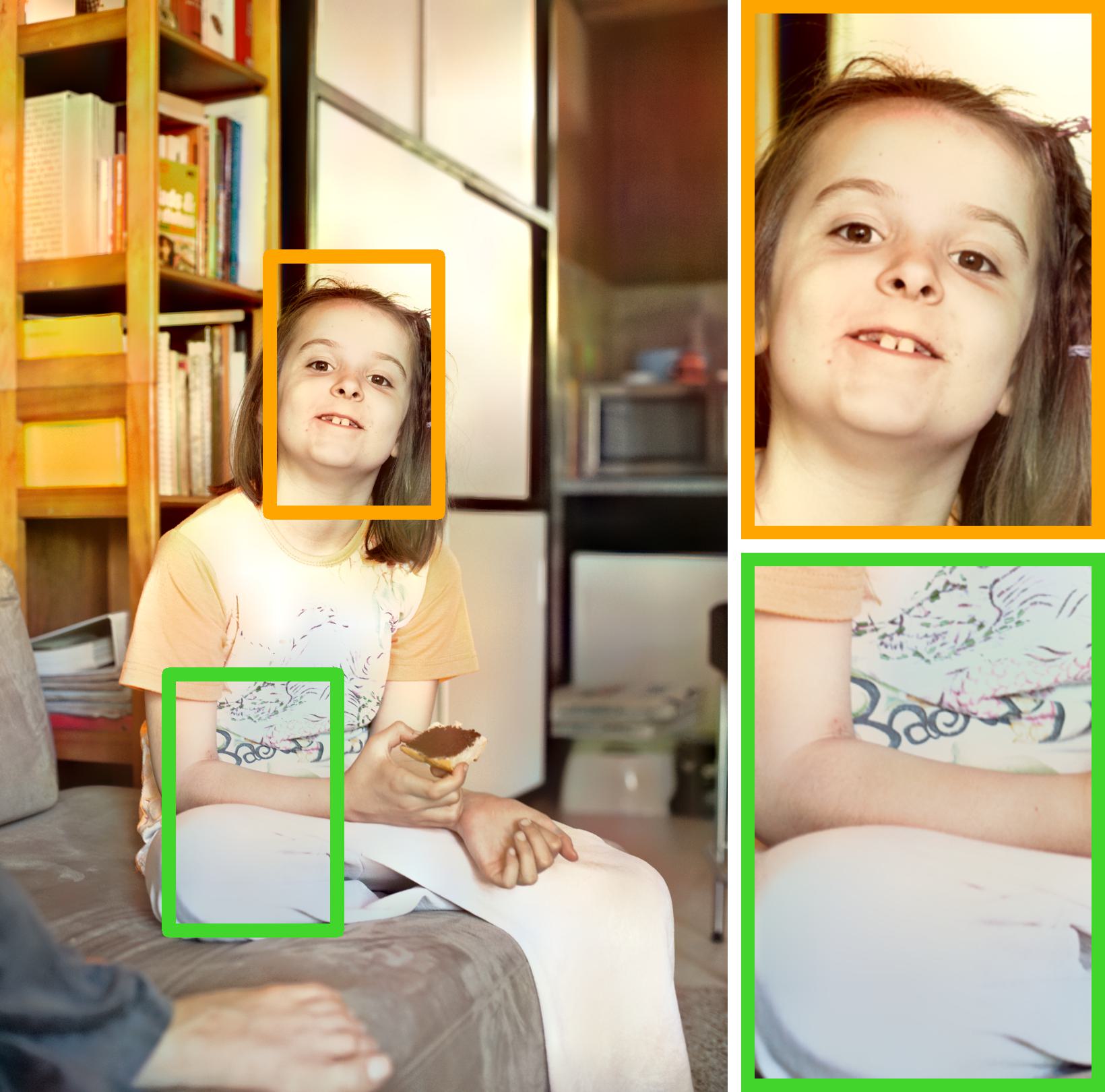} &
    \includegraphics[width=0.19\linewidth]{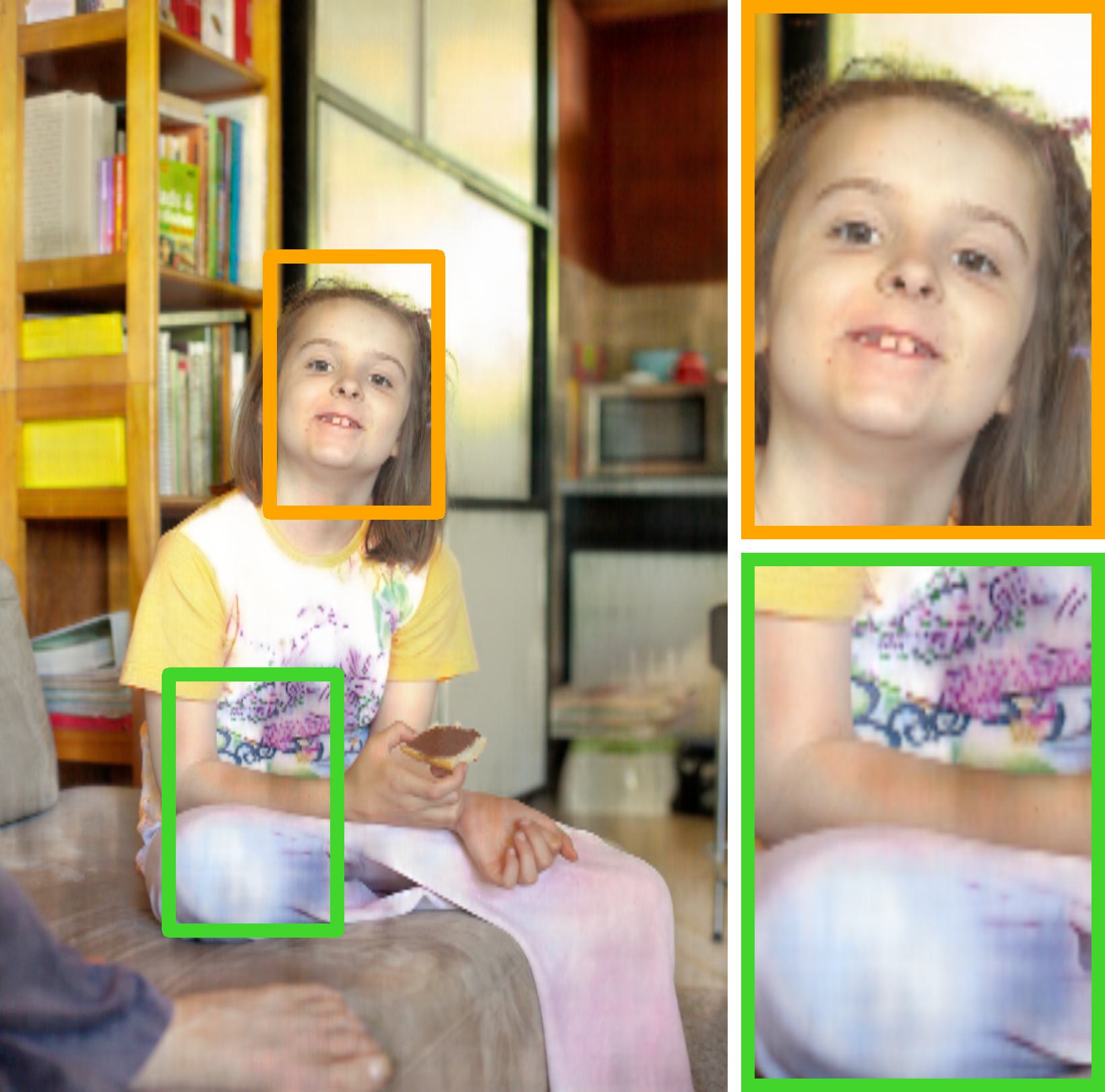} &
    \includegraphics[width=0.19\linewidth]{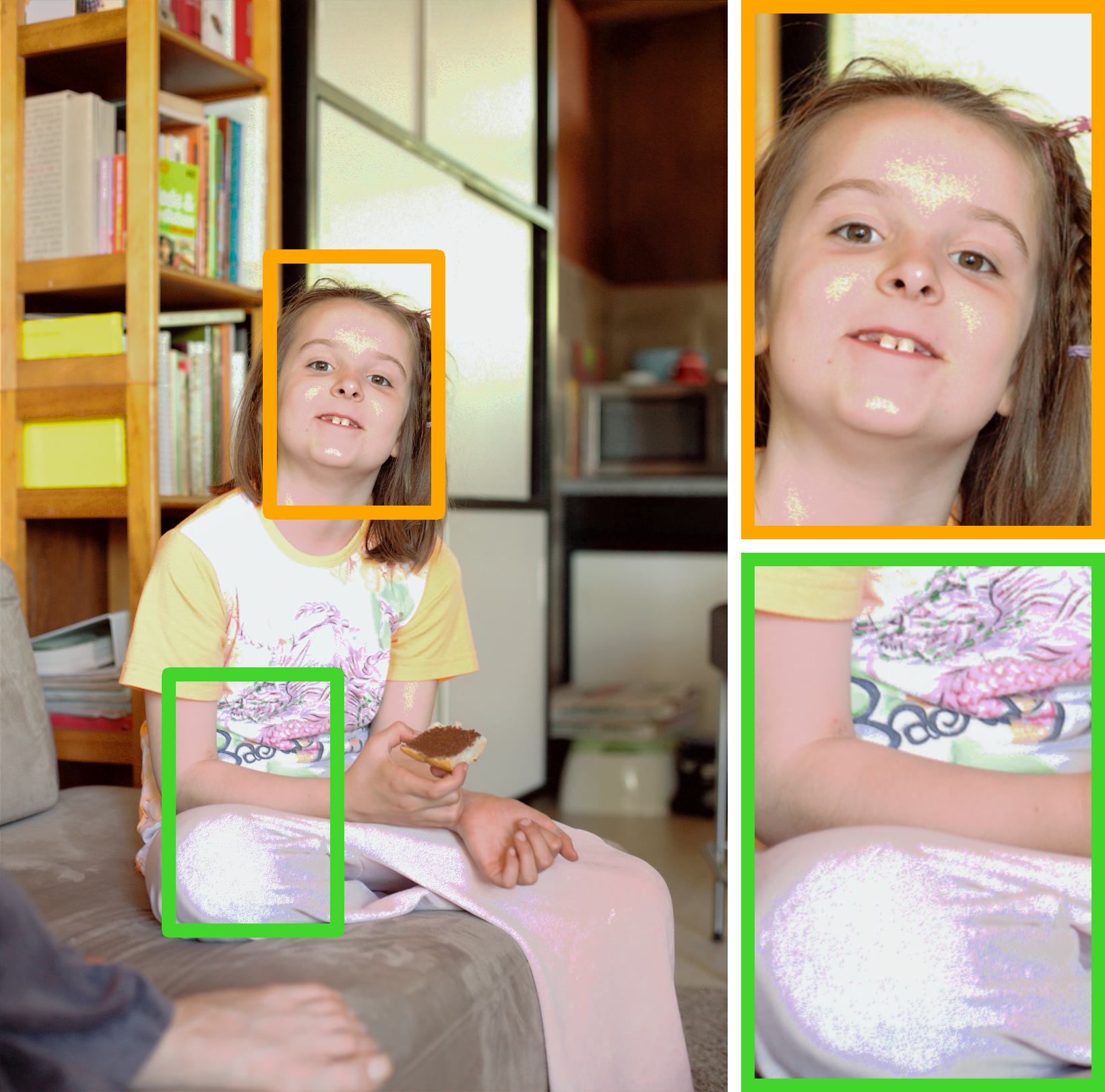}\\ 
    \scriptsize{(k) Input} & \scriptsize{(l) RetinexNet~\cite{wei_2018_BMVC_retinexnet}}
    &\scriptsize{(m) MSEC~\cite{Afifi_2021_CVPR_MSEC}} & \scriptsize{(n) SNRNet~\cite{Xu_2022_CVPR_SNR}} & \scriptsize{(o) LCDP~\cite{Wang_2022_ECCV_lcdp}}\\

    \includegraphics[width=0.19\linewidth]{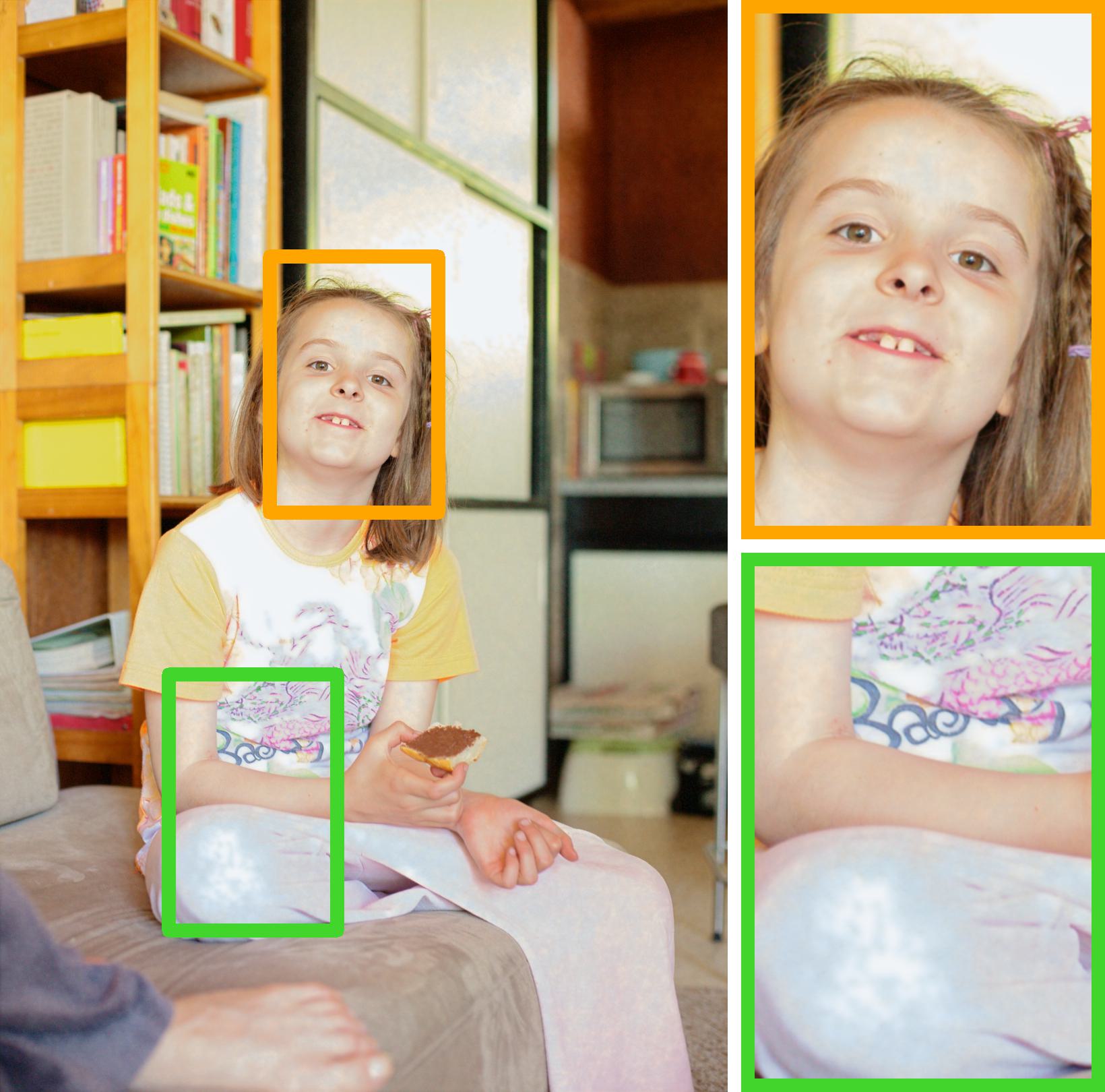} &
    \includegraphics[width=0.19\linewidth]{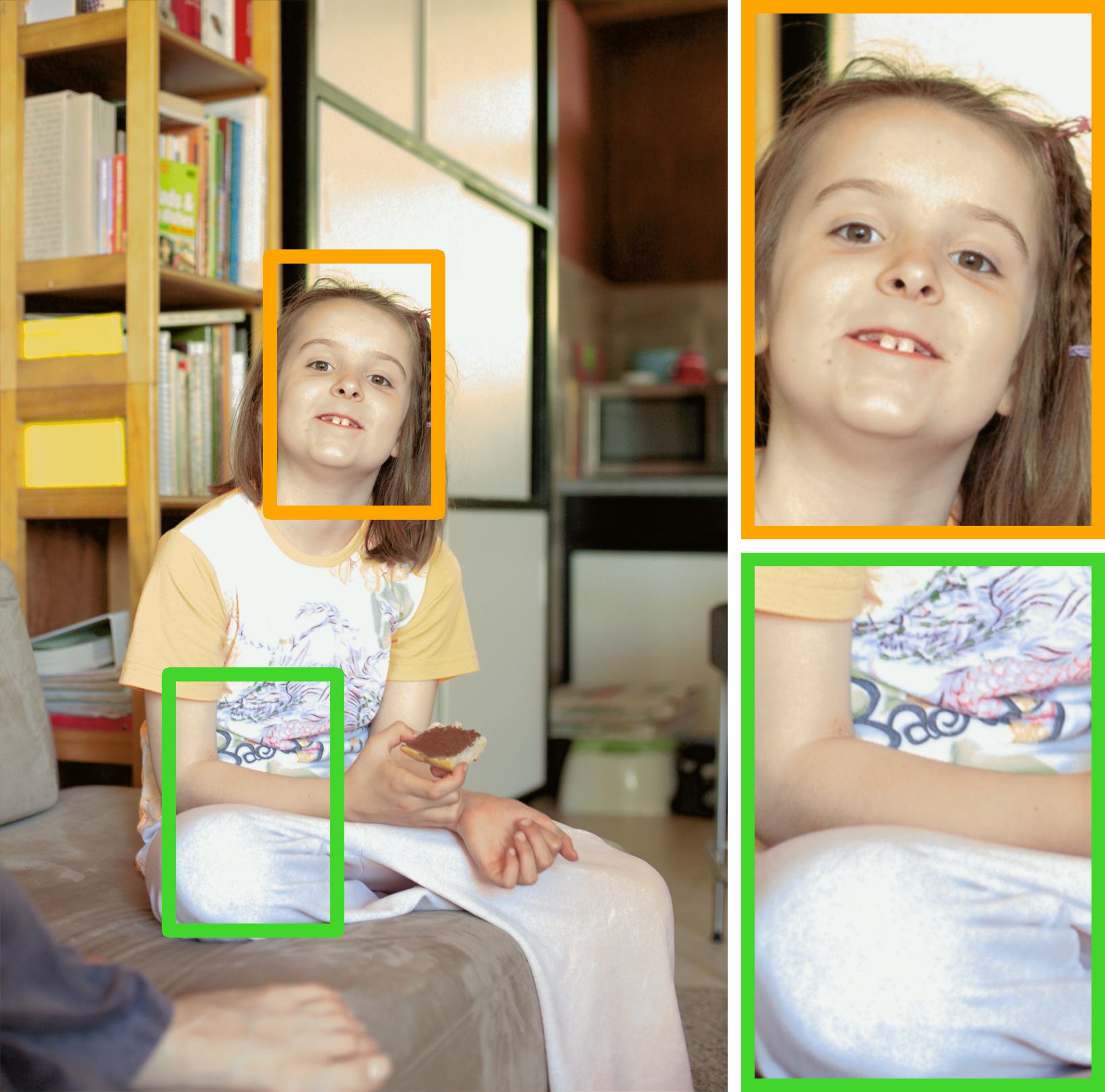} &
    \includegraphics[width=0.19\linewidth]{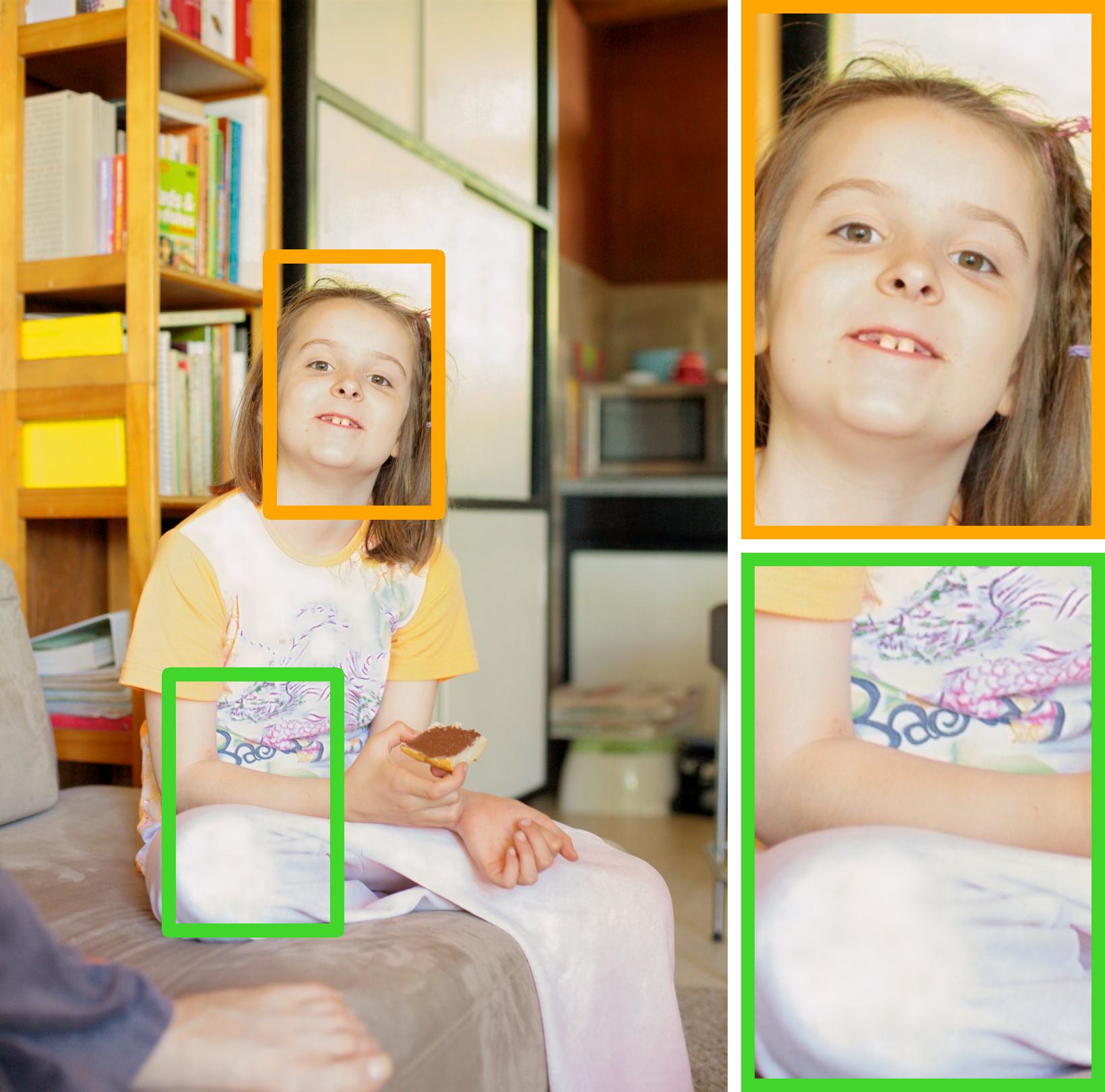} &
    \includegraphics[width=0.19\linewidth]{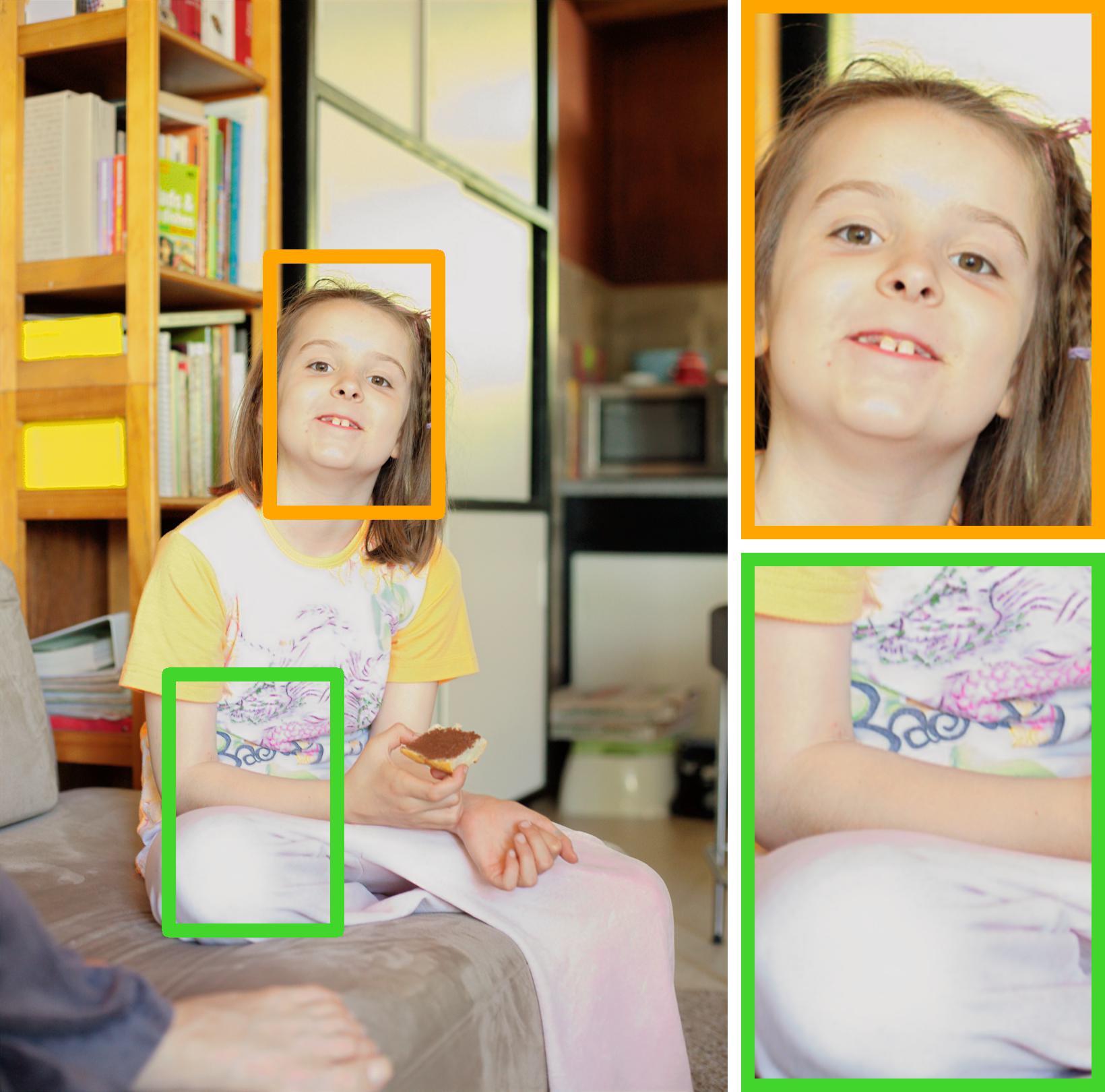} &
    \includegraphics[width=0.19\linewidth]{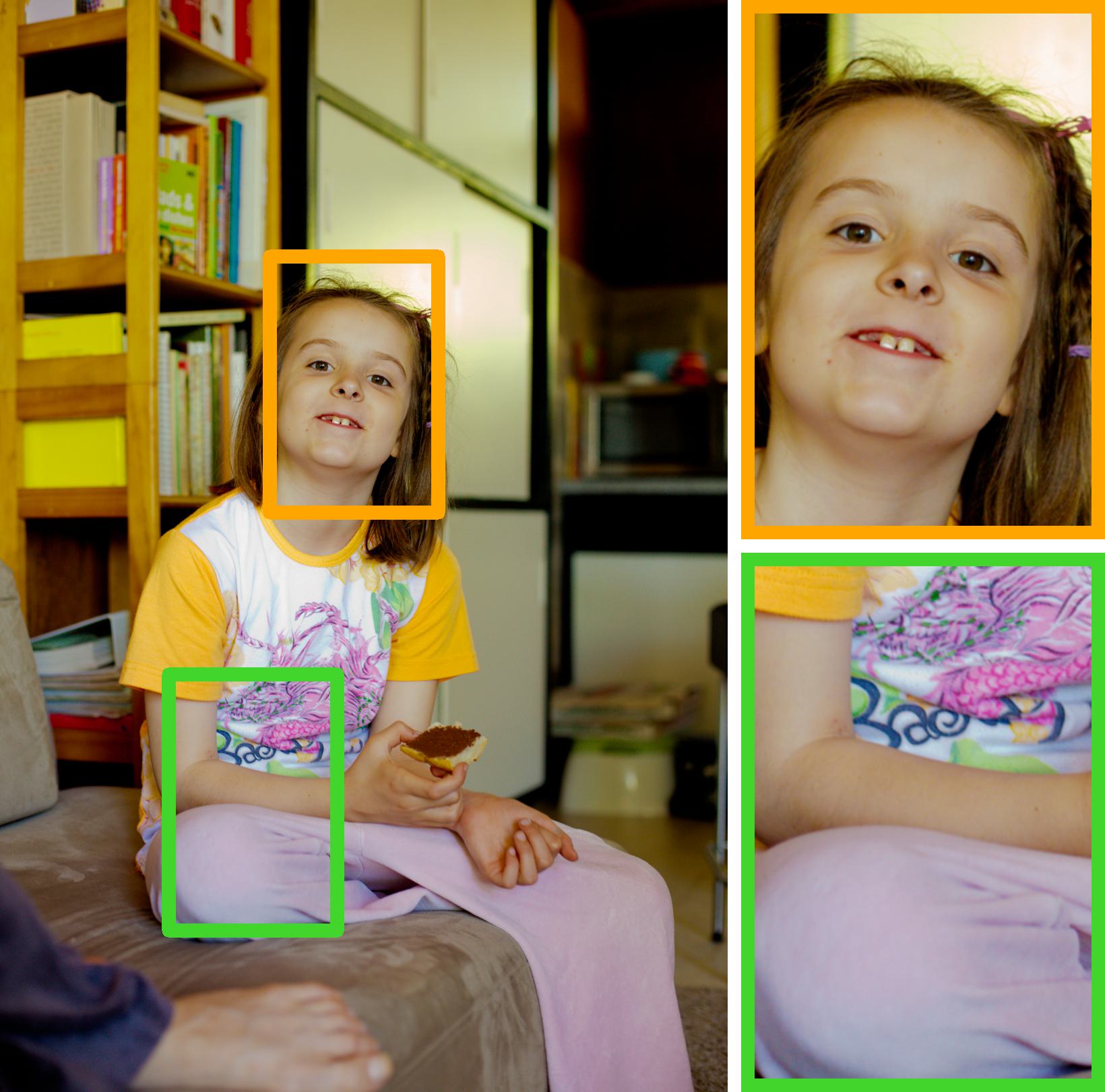}\\ 
    \scriptsize{(p) RetinexFormer~\cite{cai_2023_retinexformer}} & \scriptsize{(q) LANet~\cite{yang_2023_lanet}}
    &\scriptsize{(r) FECNet~\cite{Huang_2022_ECCV_Fourier}} & \scriptsize{(s) Ours} & \scriptsize{(t) GT}\\
    
\end{tabular}
\end{center}
\vspace{-5mm}
\caption{Visual comparison between our method and state-of-the-art methods on LCDP~\cite{Wang_2022_ECCV_lcdp} dataset, which has images with both over-exposure and under-exposure.}
\vspace{-5mm}
\label{fig:visual-comparison}
\end{figure*}

\begin{table}[t]
    \centering
    \resizebox{0.7\columnwidth}{!}{
    \begin{tabular}{l|cc}
    \hhline{===}
       Method  & PSNR~$\uparrow$ & SSIM~$\uparrow$ \\
       \hline
       HE'87~\cite{pizer_1987_histogram_equalization} & 15.873 & 0.731 \\
       MSEC'21~\cite{Afifi_2021_CVPR_MSEC} & 20.482 & 0.825 \\
       LCDPNet'22~\cite{Wang_2022_ECCV_lcdp} & 22.224 & \underline{0.849} \\
       FECNet'22~\cite{Huang_2022_ECCV_Fourier} & \underline{22.519} & 0.848 \\
       SMG'23~\cite{Xu_2023_CVPR_Structure} & 22.075 & 0.781 \\
       RetinexFormer'23~\cite{cai_2023_retinexformer} & 21.809 & 0.846 \\
       \hline
       Ours & \textbf{22.728} & \textbf{0.863} \\
    \hhline{===}
    \end{tabular}}
    \caption{Quantitative comparison between the proposed method and SOTA methods on the MSEC~\cite{Afifi_2021_CVPR_MSEC} test set. All methods are re-trained on the MSEC training set. Best performances are marked in \textbf{bold} and second best performances are \underline{underlined}.}
    \label{tab:quantitative-comparison-msec}
    \vspace{-3mm}
\end{table}

We further compare our method with LCDPNet~\cite{Wang_2022_ECCV_lcdp}, FECNet~\cite{Huang_2022_ECCV_Fourier}, SMG~\cite{Xu_2023_CVPR_Structure} and RetinexFormer~\cite{cai_2023_retinexformer}, which are the best-four performing existing methods according to Table~\ref{tab:quantitative-comparison-lcdp}, on the MSEC~\cite{Afifi_2021_CVPR_MSEC} dataset. We also report the performance of MSEC~\cite{Afifi_2021_CVPR_MSEC} and HE~\cite{pizer_1987_histogram_equalization} for reference. 
As shown in Table~\ref{tab:quantitative-comparison-msec}, our method also outperforms these existing methods, which demonstrates that our color shift estimation and correction method can handle either over- or under-exposure at image level well.

\noindent{\bf Qualitative Comparisons.} We further visually compare the results of the proposed method and state-of-the-art methods in Figure~\ref{fig:visual-comparison}, where input images are from the LCDP~\cite{Wang_2022_ECCV_lcdp} dataset. We can see that methods proposed for handling either over- or under-exposures (\ie, MSEC~\cite{Afifi_2021_CVPR_MSEC}, LANet~\cite{yang_2023_lanet}, and FECNet~\cite{Huang_2022_ECCV_Fourier}) cannot rectify the brightness of over- and under-exposed regions at the same time. For example, they either fail to restore the greenish color of under-exposed grasses, or fail to restore the bluish color of the over-exposed sky (see Figure~\ref{fig:visual-comparison} (c, g, and h)).
Methods originally proposed for low-light image enhancement (\ie, RetinexNet~\cite{wei_2018_BMVC_retinexnet}, SNRNet~\cite{Xu_2022_CVPR_SNR}, and RetinexFormer~\cite{cai_2023_retinexformer}) tend to increase the overall image brightness, which further buries the details of clouds (Figure~\ref{fig:visual-comparison} (b, d, and f)).
LCDP~\cite{Wang_2022_ECCV_lcdp} is designed to deal with the coexistence of over- and under-exposure; however, the sky regions after the enhancement tend to have purplish colors (e).
Similar problems can be found in the second case. In contrast, our method can enhance these images with better details and colors (Figure~\ref{fig:visual-comparison} (i and s)).

\begin{figure}[h]
\renewcommand{\tabcolsep}{0.8pt}
\begin{center}

\begin{tabular}{cccc}

    \includegraphics[width=0.24\linewidth]{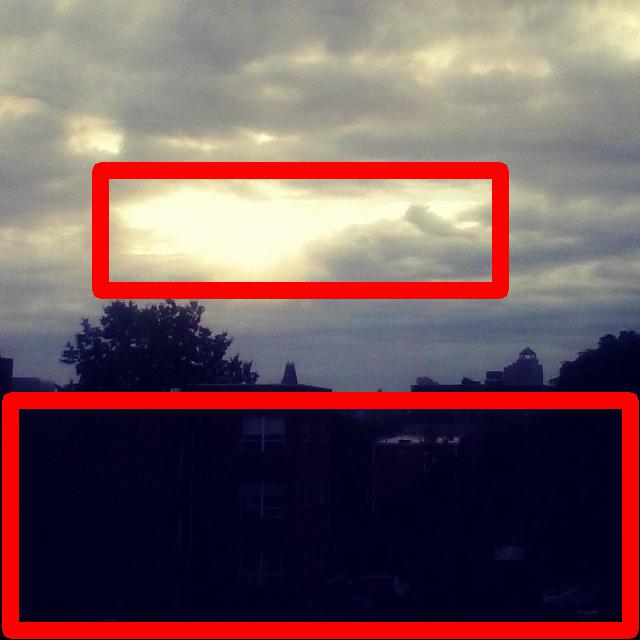} &
    \includegraphics[width=0.24\linewidth]{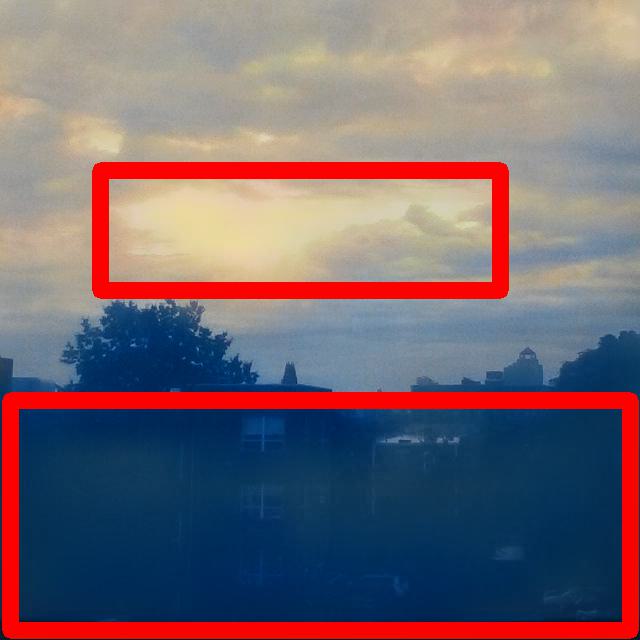} &
    \includegraphics[width=0.24\linewidth]{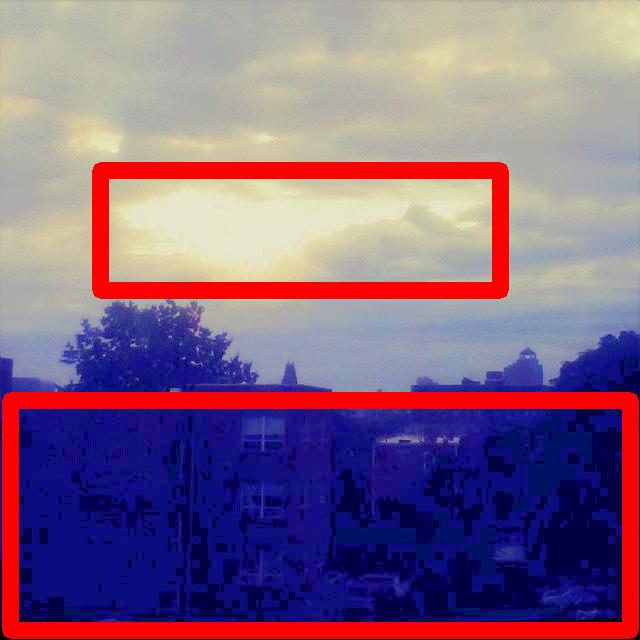} &
    \includegraphics[width=0.24\linewidth]{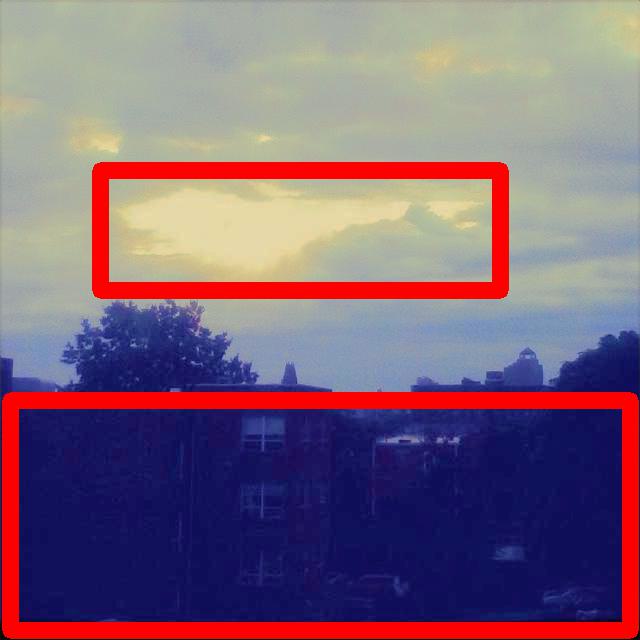}\\



    \includegraphics[width=0.24\linewidth]{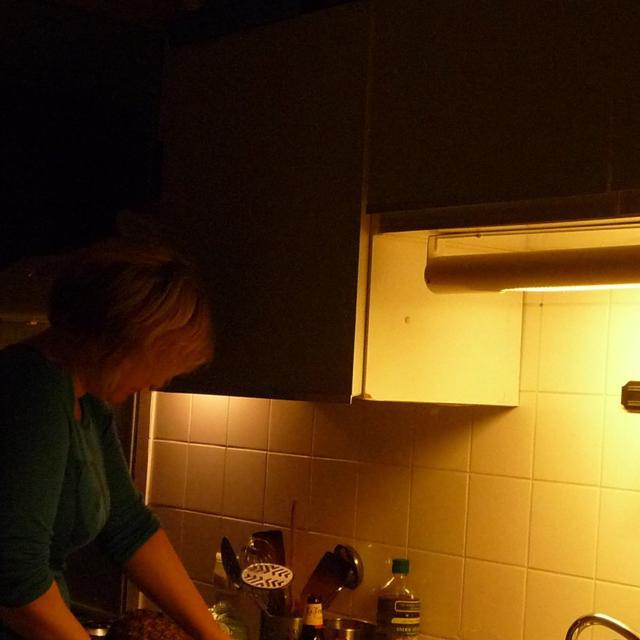} &
    \includegraphics[width=0.24\linewidth]{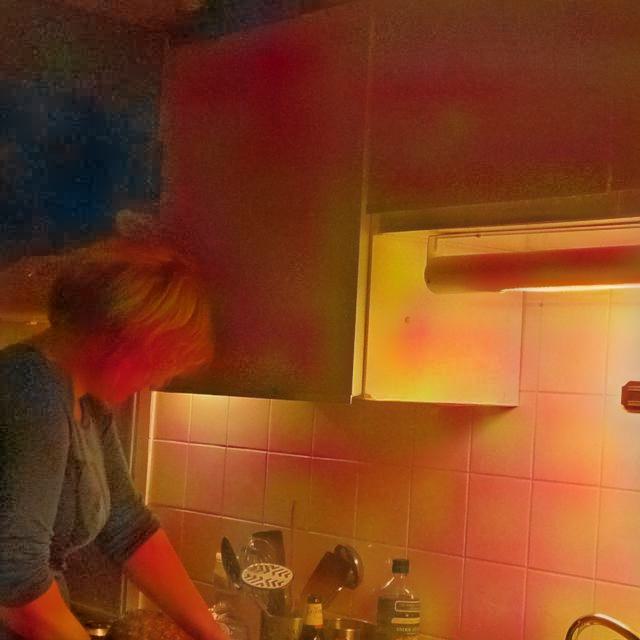} &
    \includegraphics[width=0.24\linewidth]{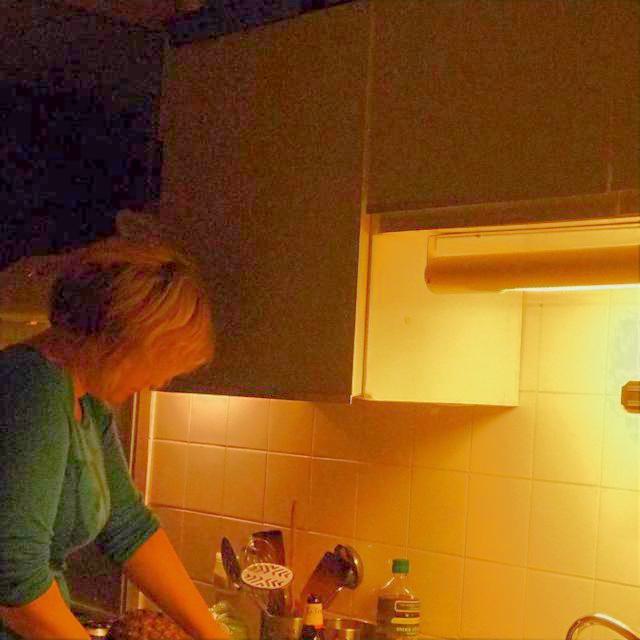} &
    \includegraphics[width=0.24\linewidth]{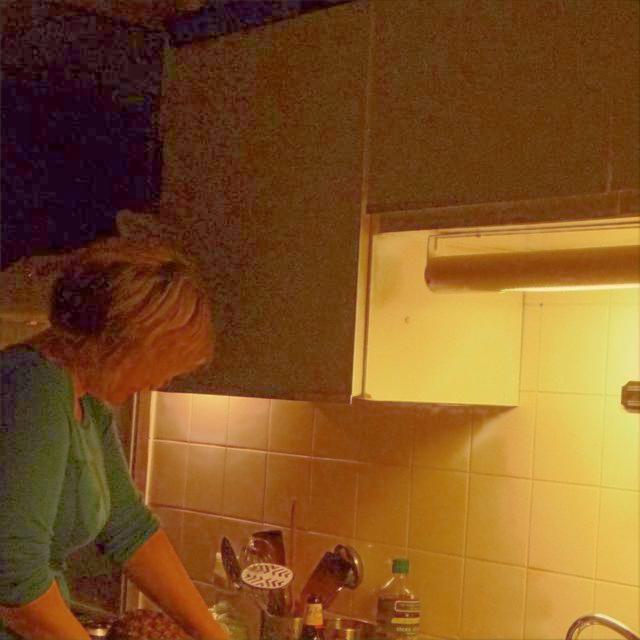}\\

    \includegraphics[width=0.24\linewidth]{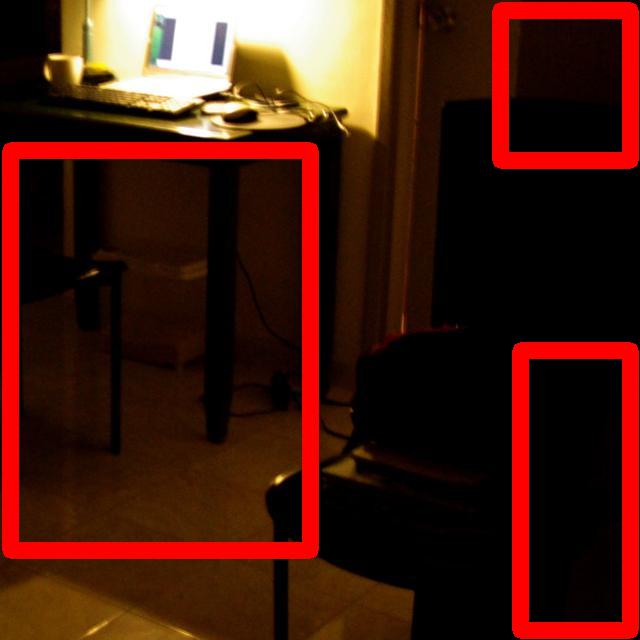} &
    \includegraphics[width=0.24\linewidth]{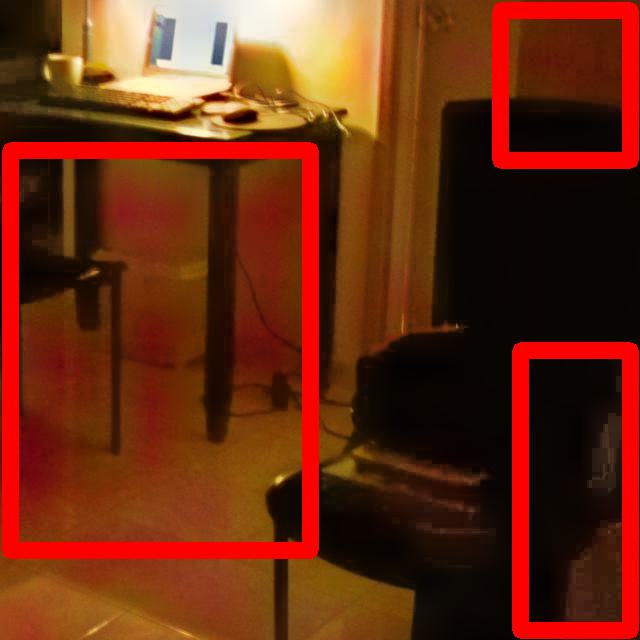} &
    \includegraphics[width=0.24\linewidth]{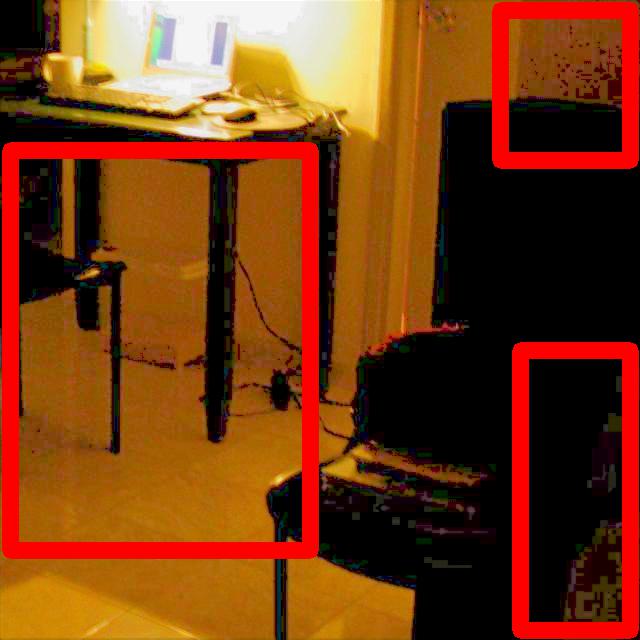} &
    \includegraphics[width=0.24\linewidth]{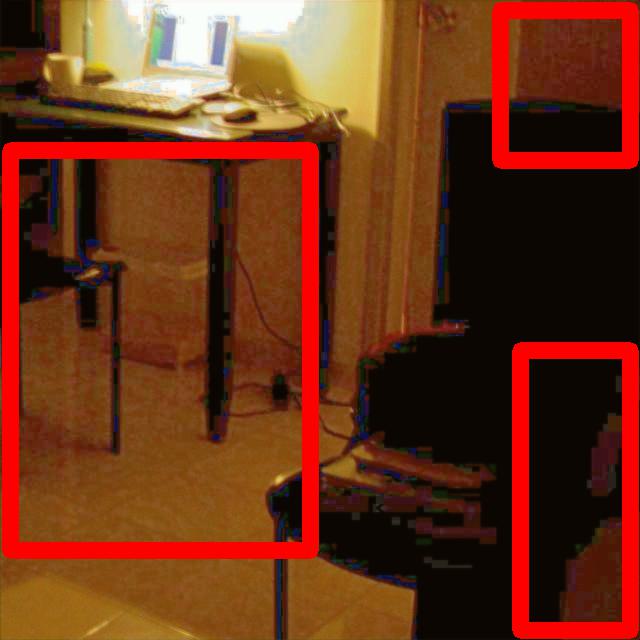}\\

    \includegraphics[width=0.24\linewidth]{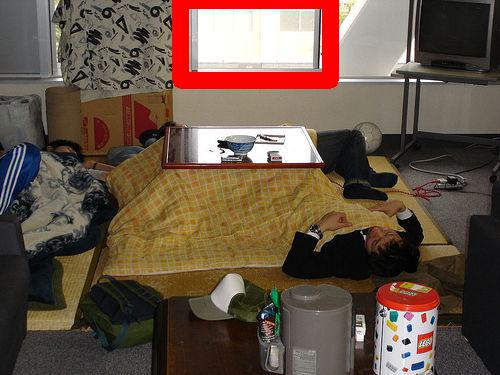} &
    \includegraphics[width=0.24\linewidth]{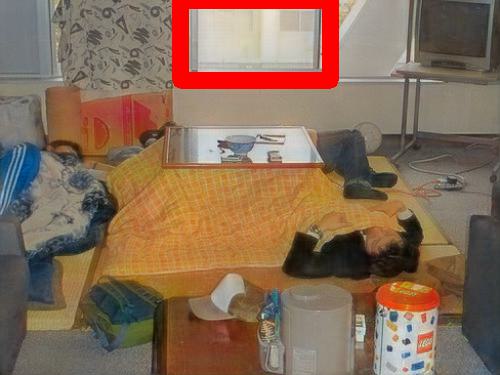} &
    \includegraphics[width=0.24\linewidth]{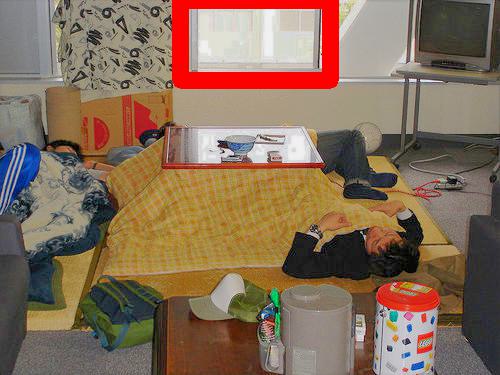} &
    \includegraphics[width=0.24\linewidth]{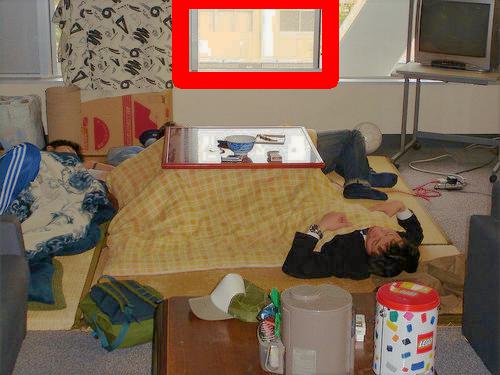}\\

    \includegraphics[width=0.24\linewidth]{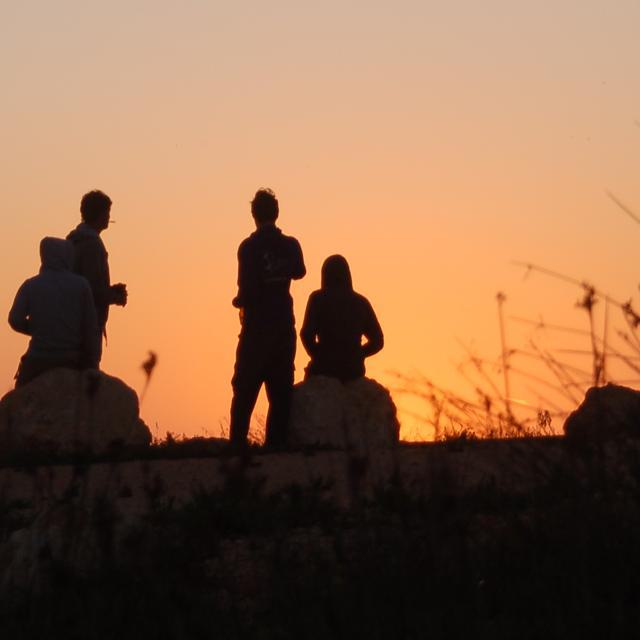} &
    \includegraphics[width=0.24\linewidth]{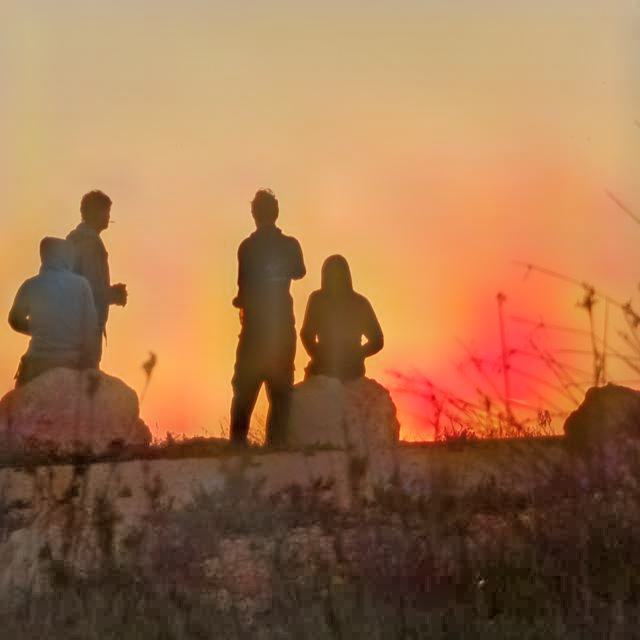} &
    \includegraphics[width=0.24\linewidth]{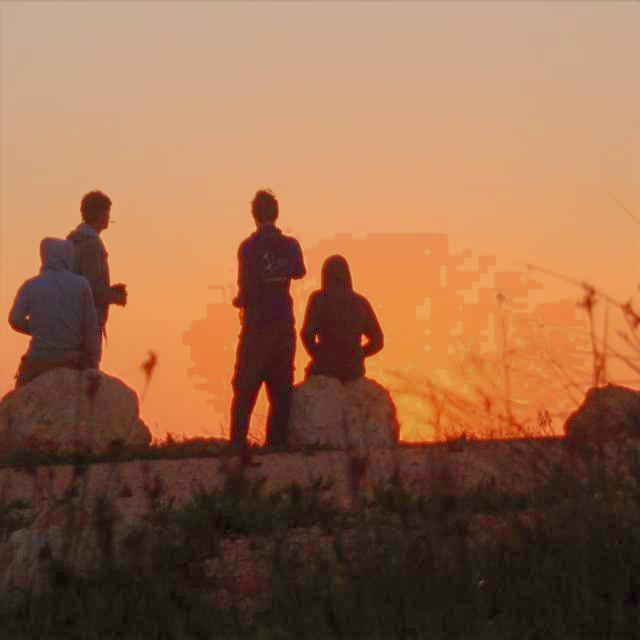} &
    \includegraphics[width=0.24\linewidth]{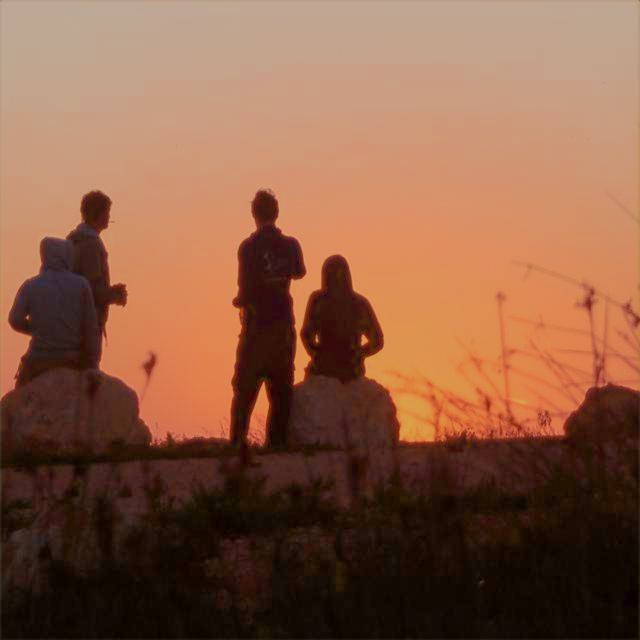}\\
    
    \scriptsize{Input} & \scriptsize{MSEC~\cite{Afifi_2021_CVPR_MSEC}}
    &\scriptsize{LCDP~\cite{Wang_2022_ECCV_lcdp}} & \scriptsize{Ours}\\
    
\end{tabular}
\end{center}
\vspace{-5mm}
\caption{Visual results of real-world examples from the Flickr 30K~\cite{flickr30k} dataset. Our method produces better visual results.}
\vspace{-5mm}
\label{fig:real-world}
\end{figure}

Figure~\ref{fig:real-world} further shows visual results of real-world images from the Flickr 30K~\cite{flickr30k} dataset. While existing methods tend to produce obvious color artifacts after enhancements, our method produces results of better visual quality.

\subsection{Ablation Study}

We now conduct ablation studies on the proposed network designs in Table~\ref{tab:ablation-network}.
First, we replace the proposed color space deformable convolution with \yycr{three different types of} spatial deformable convolution\yycr{s} \yycr{(\ie, deformable convolution with only spatial offsets $Ours_{Spatial}$, deformable convolution with spatial offsets and modulation scalars $Ours_{Spatial+Modul}$, and deformable convolution with spatial offsets and proposed color offsets $Ours_{Spatial+Color}$)}, and with the standard convolution (denoted as $Ours_{Conv}$), to test the effectiveness of the proposed Color Shift Estimation module.
Second, we replace the proposed Color Modulation module with the original non-local block and concatenate $O_B$, $O_D$, and $I_x$ 
as input (denoted as $Ours_{NL}$) to test the effectiveness of Color Modulation module.
The first three rows in Table~\ref{tab:ablation-network} show that replacing either the proposed COSE or COMO modules results in the degradation of the enhancement performance.

We also investigate different pipeline variants. First, we change the over-exposed illumination map $F_L^O$ from $f(1-I_x)$ to $1 - f(I_X)$ (where $f(\cdot)$ denotes the UNet-based feature extractor), to make a strict opposition between over-exposed map and under-exposed map (denoted as $Ours_{opposition}$).
Second, we change two illumination maps from having 1-channel to 3-channels (which is adopted in LCDP~\cite{Wang_2022_ECCV_lcdp} and DeepUPE~\cite{wang_2019_CVPR_deepupe}) (denoted as $Ours_{3channel}$).
We also use an independent pseudo-normal feature generator that does not share parameters with the color modulation module (denoted as $Ours_{NoShare}$).
The 4th to 6th rows in Table~\ref{tab:ablation-network} show that using other pipeline variants may significantly degrade the performance, which verifies the effectiveness of our designs.

Last, we perform ablation studies on the loss function. Specifically, we first remove $\mathcal{L}_{pseudo}$ and only supervise the final result (denoted as $Ours_{output}$). Second, we remove the VGG loss (denoted as $Ours_{w/oVGG}$). Last, we remove both SSIM loss and VGG loss (denoted as $Ours_{w/oVGGSSIM}$). The 7th to 9th rows shows that adding these losses help improve the performance of our network.

\begin{table}[t]
    \centering
    \resizebox{0.6\columnwidth}{!}{
    \begin{tabular}{l|cc}
    \hhline{===}
       Options  & PSNR~$\uparrow$ & SSIM~$\uparrow$ \\
       \hline
       $Ours_{Conv}$ & 23.443 & 0.834 \\
       $Ours_{Spatial}$ & 22.623 & 0.827 \\
       $Ours_{Spatial+Modul}$ & 23.569 & 0.840 \\
       $Ours_{Spatial+Color}$ & 23.415 & 0.836 \\
       \hline
       $Ours_{NL}$ & 23.201 & 0.844 \\
       $Ours_{opposition}$ & 21.829 & 0.799 \\
       $Ours_{3channel}$ & 23.016 & 0.830 \\
       $Ours_{NoShare}$ & 23.513 & 0.844 \\
       $Ours_{output}$ & 23.588 & 0.836 \\
       $Ours_{w/oVGG}$ & 23.593 & 0.846 \\
       $Ours_{w/oVGGSSIM}$ & 23.512 & 0.837 \\
       \hline
       $Ours$ & \textbf{23.627} & \textbf{0.855} \\
    \hhline{===}
    \end{tabular}}
    \vspace{-2mm}
    \caption{Ablation study. Best results are marked in {\bf bold}.}
    \label{tab:ablation-network}
    \vspace{-3mm}
\end{table}


%% file: sec/6_conclusion.tex
\section{Conclusion}
\label{conclusion}
In this paper, we have studied the problem of enhancing images with both over- and under-exposures.
We have observed opposite color tone distribution shifts between over- and under-exposed pixels, and a lack of ``normal-exposed'' regions/pixels as reference resulting in the unsatisfactory performance of existing methods.
We have proposed a novel method to enhance images with both over- and under-exposures by learning to estimate and correct the color tone shifts.
Our method has a novel color shift estimation module to estimate the color shifts based on our created pseudo-normal color feature maps and correct the estimated color shifts of the over- and under-exposed regions separately.
A novel color modulation module is proposed to modulate the separately corrected colors in the over- and under-exposed regions to produce the enhanced image.
We have conducted extensive experiments to show that our method outperforms existing image enhancement approaches.

\begin{figure}[h]
\renewcommand{\tabcolsep}{0.8pt}
\begin{center}

\begin{tabular}{cc}

    \includegraphics[width=0.48\linewidth]{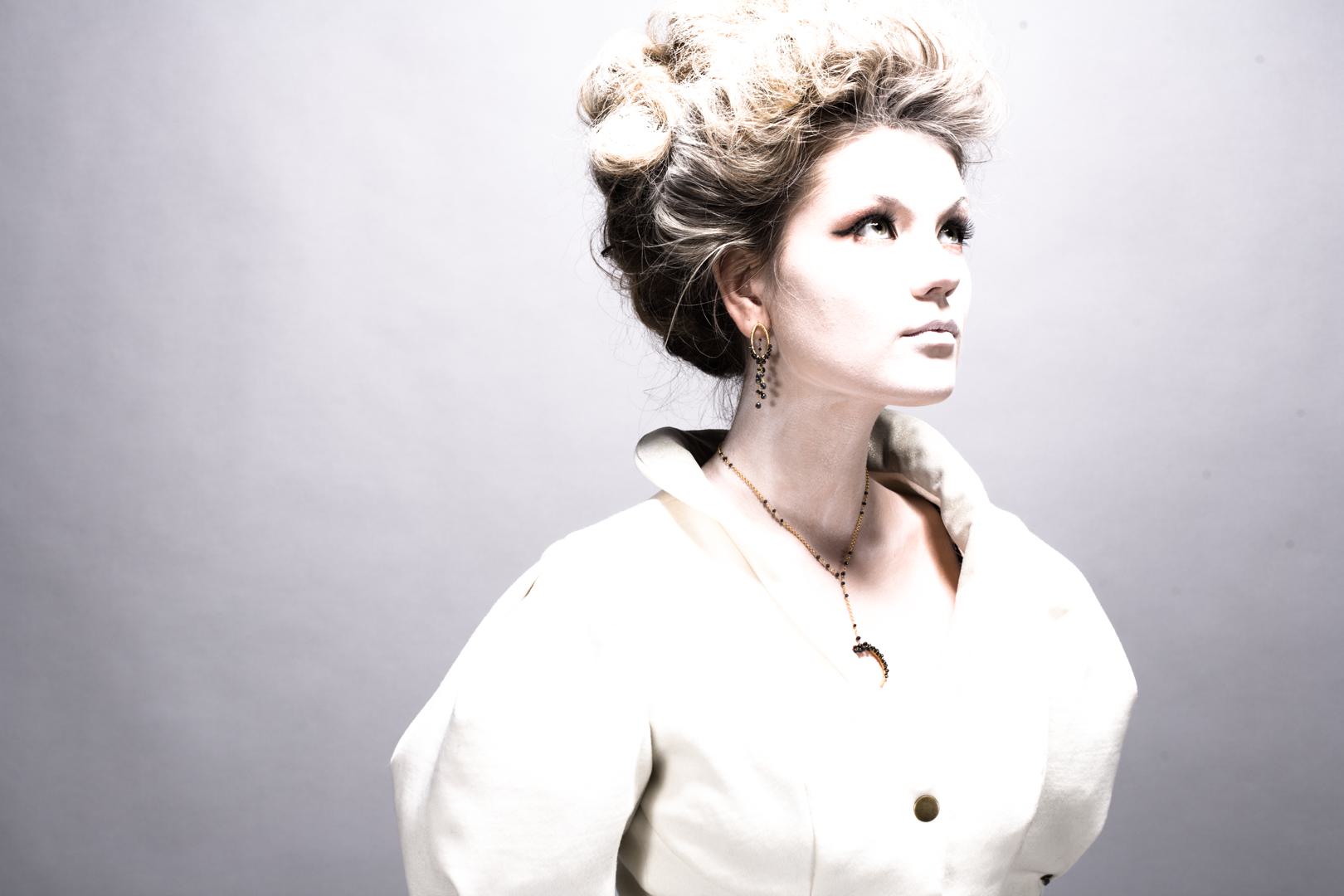} &
    \includegraphics[width=0.48\linewidth]{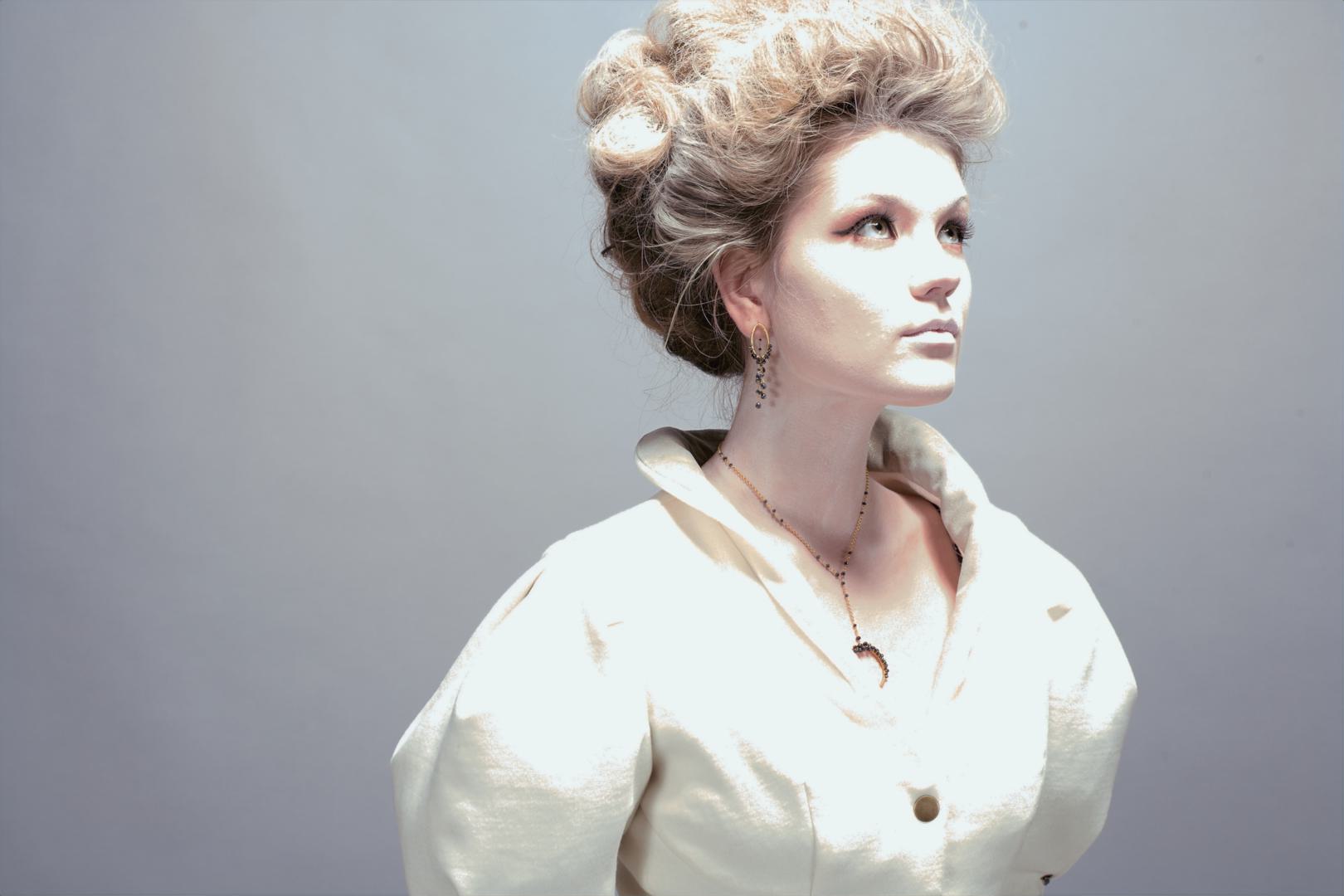} \\ 
    \scriptsize{(a) Input} & \scriptsize{(b) Ours} \\

    
\end{tabular}
\end{center}
\vspace{-5mm}
\caption{Our method may fail when pixels are completely saturated \ryn{in over-exposed regions (\eg, the forehead and cheeks)} (a), where there is insufficient contextual information for our model to estimate color shifts and restore the original colors (b).}
\vspace{-5mm}
\label{fig:limitations}
\end{figure}

\kk{Our method does have limitations.}
%
Since our method relies on the generation of pseudo-normal feature maps to help estimate and correct the brightness and color shifts, it may fail in scenarios where the over-exposed pixels are completely saturated (\eg, the forehead and cheeks in Figure~\ref{fig:limitations}(a)), resulting in an unsatisfactory enhancement result (Figure~\ref{fig:limitations}(b)). Incorporating generative models into our pseudo-normal feature generator may help address such limitations and can be interesting for future work. \\

\noindent {\bf Acknowledgements.} This work is partly supported by an ITF grant from the Innovation and Technology Commission of Hong Kong SAR (ITC Ref.: PRP/003/22FX).

